\documentclass[runningheads]{llncs}
\usepackage{graphicx}
\usepackage{pythonhighlight}
\usepackage{array}

\usepackage{tikz}
\usepackage{comment}
\usepackage{amsmath,amssymb} 
\usepackage{color}
\usepackage{breakcites}

\usepackage{wrapfig} 
\usepackage{multirow} 
\usepackage{booktabs} 

\usepackage[accsupp]{axessibility}  

\newcommand{\etal}{\emph{et al}.}

\begin{document}
\pagestyle{headings}
\mainmatter
\def\ECCVSubNumber{1531}  

\title{MIME: Minority Inclusion for Majority Group Enhancement of AI Performance} 

\titlerunning{Minority Inclusion for Majority Enhancement of AI Performance}
%

\author{Pradyumna Chari\inst{1}
\and
Yunhao Ba\inst{1}
\and
Shreeram Athreya\inst{1}
\and
Achuta Kadambi\inst{1,2}
}

\authorrunning{P. Chari et al.}
%
\institute{Department of Electrical and Computer Engineering, UCLA\\ \and
Department of Computer Science, UCLA\\
\email{\{pradyumnac,yhba,shreeram\}@ucla.edu}, \email{achuta@ee.ucla.edu}}
\maketitle

\begin{abstract}
Several papers have rightly included minority groups in artificial intelligence (AI) training data to improve test inference for minority groups and/or society-at-large. A society-at-large consists of both minority and majority stakeholders. A common misconception is that minority inclusion does not increase performance for majority groups alone. In this paper, we make the surprising finding that including minority samples can improve test error for the majority group. In other words, minority group inclusion leads to majority group enhancements (MIME) in performance. A theoretical existence proof of the MIME effect is presented and found to be consistent with experimental results on six different datasets. Project webpage: \url{https://visual.ee.ucla.edu/mime.htm/}. 
\end{abstract}

\section{Introduction}
Inclusion of minorities in a dataset impacts the performance of artificial intelligence (AI). Recent research has  presented the value of inclusive datasets to improve AI performance on minorities and also for society-at-large~\cite{gebru2018datasheets,buolamwini2018gender,larrazabal2020gender,ryu2017inclusivefacenet,li2019repair,mehrabi2021survey,jo2020lessons,gong2019diversity,kadambi2021achieving}. A society-at-large consists of both majority and minority stakeholders. However, an objection (often silently posed) to minority inclusion efforts, is that the inclusion of minorities can diminish performance for the majority. This is based on a ``rule of thumb'' that AI performance is maximized when one trains and tests on the same distribution. A devil's advocate position against minority inclusion might be presented as: ``In a fictitious society where we are absolutely certain that only blue-skinned humans will exist in the test set, why include out of distribution orange-skinned humans in the training set?". 

\begin{figure*}[t]
    \centering
    \includegraphics[width=0.65\textwidth]{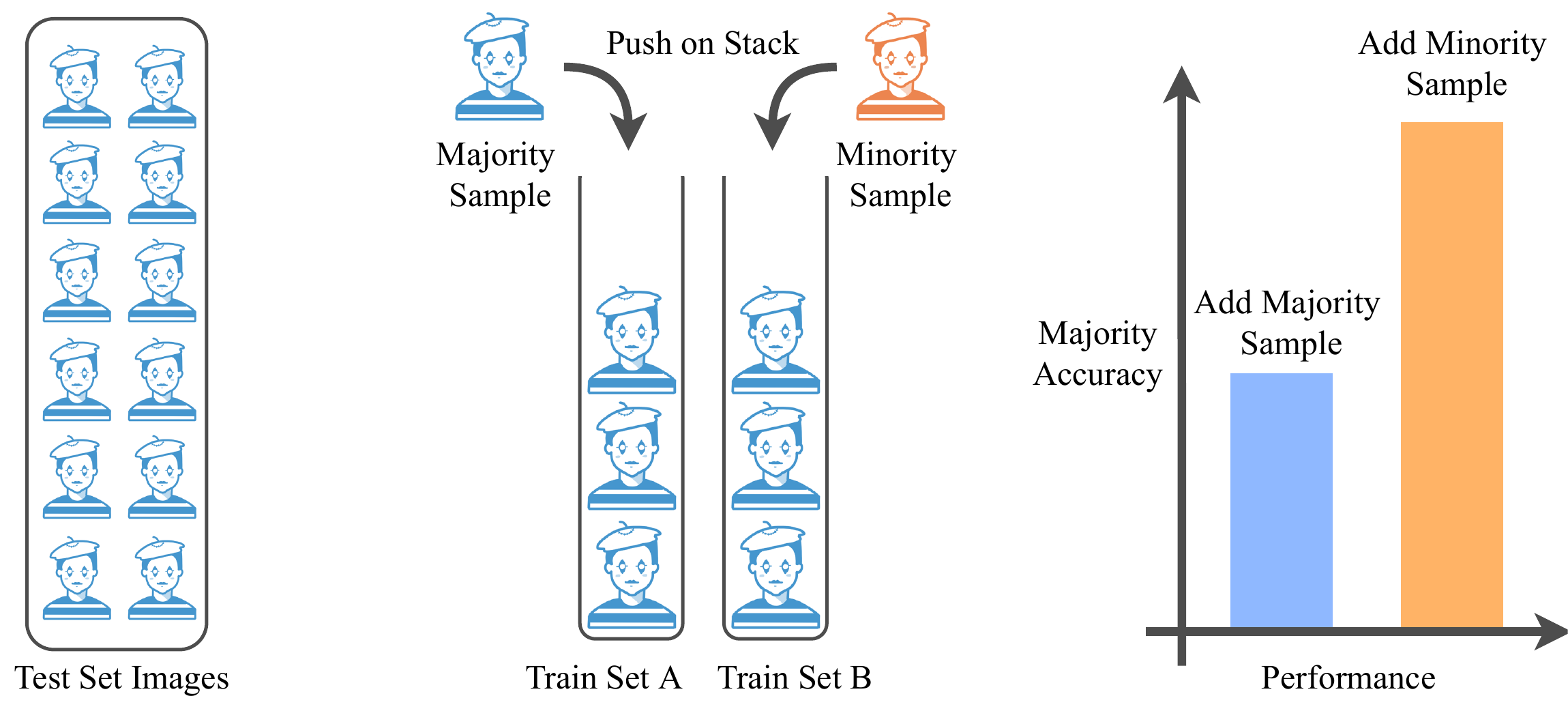} 
     \caption{\textbf{Inclusion of minorities can improve performance for majorities.} We theoretically describe an effect called Minority Inclusion, Majority Enhancement (MIME). The figure depicts test classification of blue mimes, and an initial training stack, also of blue mimes. If allowed to add one more training sample, it can be better to push an orange mime onto the training stack rather than a blue mime. Test accuracy can increase by pushing orange, even though the test set consists of blue mimes alone.}
    \label{fig:teaser}
\end{figure*}

In this paper, we make the surprising finding that inclusion of minority samples improves AI performance not just for minorities, not just for society-at-large, but \emph{even for majorities}. We refer to this effect as Minority Inclusion, Majority Enhancement (MIME), illustrated in Figure~\ref{fig:teaser}. Specifically, we note that including some minority samples in the train set improves majority group test performance. However, continued addition of minority samples leads to performance drop. The effect holds under statistical conditions that are represented in traditional computer vision datasets including FairFace~\cite{karkkainen2021fairface}, UTKFace~\cite{zhang2017age}, pets~\cite{golle2008machine}, medical imaging datasets~\cite{rajpurkar2017chexnet} and even non-vision data~\cite{blake1998uci}. Although deep learning is used for these problems, the flattening layer of a network can be empirically approximated to elementary distributions like Gaussian Mixture Models (GMMs). A GMM facilitates closed-form analysis to prove the existence of the MIME effect. Additionally, we show existence of MIME on general distributions. Classification experiments on neural networks validate using Gaussian mixtures: complex neural networks exhibit feature embeddings in flat layers, distributed with approximately Gaussian density, across six datasets, in and beyond computer vision, and across many realizations and configurations. 

\begin{figure*}[t]
    \centering
    \includegraphics[width=0.5\textwidth]{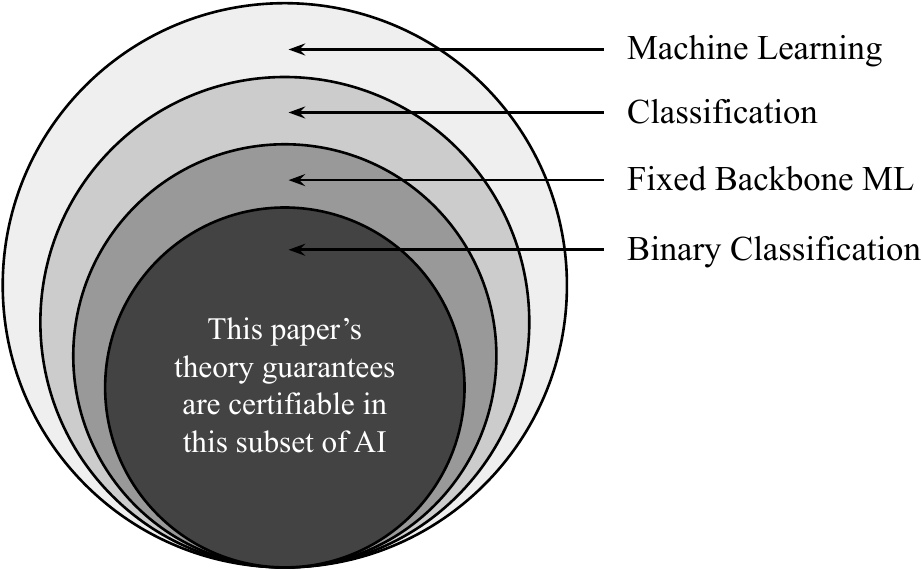}
     \caption{\textbf{This paper proves* that including minorities improves majority performance.} *When do the provable guarantees hold? The guarantees are certifiable for fixed backbone binary classification (e.g. one uses a head network with pretrained weights and fine-tunes a downstream layer for classification). The fixed backbone ML is far from a toy scenario (it is considered SoTA by some authors~\cite{kang2019decoupling}) and also enables provable certification - ordinarily it is hard to prove things for neural network settings.}
    \label{fig:theory_scope}
\end{figure*}

Fairness in machine learning is an exceedingly popular area, and our results benefit from several key papers published in recent years. Sample reweighting approaches recognize the need to preferentially weight difficult examples~\cite{dong2017class,ren2018learning,cui2019class}. Active and online learning benefit from insights into sample ``informativeness'' (i.e. given a budget on the number of training samples, which would be the best sample to include~\cite{choi2021active,dasgupta2011two}). Domain randomization literature indicates that surprising perturbations to the training set can improve generalization performance~\cite{tremblay2018training,yue2019domain,huang2021fsdr}. We extend some of these theoretical insights to the sphere of analyzing benefits of minority inclusion on majority performance.

\subsection{Contributions}
\label{sec:contributions}

\noindent While some works~\cite{gwilliam2021rethinking,larrazabal2020gender} have observed related phenomena for isolated tasks, to the best of our knowledge, characterizing benefits to majority groups by including minority data is largely unexplored theoretically. Our contributions are as follows:
\begin{itemize}
    \item We introduce the Minority Inclusion Majority Enhancement (MIME) effect in a theoretical and empirical setting.
    \item Theoretically: we derive in closed form, the existence of the MIME effect both with and without domain gap (Key Results 1 and 2) and for general sample distributions (Key Result 3). 
    \item Empirically: we test the MIME effect on six datasets, as varied as animals to medical images, and observe the existence of MIME consistent with theory.
\end{itemize}

\subsection{Outline of Theoretical Scope}
\label{sec:theory_scope}
\noindent Figure~\ref{fig:theory_scope} describes the theoretical scope. Through three key results (Theorem 1, Theorem 2 and Theorem 3), this paper offers an existence proof of the MIME effect. An existence proof can leverage a tractable setting. As in Figure~\ref{fig:teaser}, training data is a stack of $K-1$ majority samples. Test data is all majority samples. We can push one additional training sample to increase the stack size to $K$. We are allowed the choice of having the $K$-th sample drawn from the minority or majority group. Theorem 1 proves that, under the assumptions in Section~\ref{sec:math}, pushing a minority sample is superior for majority group performance improvements. Theorem 2 generalizes this result to a more realistic scenario, with domain gap. Theorem 3 extends the existence proof to general sample distributions. Empirical results on real-world AI tasks offer validation for theoretical assumptions.
\label{sec:intro}

\section{Related Work}

\noindent\textbf{Debiasing and fairness: }
It has been widely reported that biases in training data lead to biased algorithmic performance~\cite{bolukbasi2016man,hendricks2018women,buolamwini2018gender}. 
Work has been carried out in identifying and quantifying biases~\cite{balakrishnan2021towards,bellamy2019ai,wang2019balanced} and a range of methods exist to address them~\cite{gong2019diversity,mehrabi2021survey}. Early approaches suggest oversampling strategies~\cite{elkan2001foundations,bickel2009discriminative}. Other methods propose resampling based on individual performance~\cite{li2019repair}. Some works utilize information bottlenecks to disentangle biased attributes~\cite{tartaglione2021end}. Still other methods propose bias mitigation solutions based on adversarial learning~\cite{zhang2018mitigating} or include considerations like protected class-specific classifiers~\cite{wang2020towards}. Generative models have also found use in creating synthetic datasets with debiased attributes~\cite{ramaswamy2021fair}. Xu \etal~\cite{xu2021robust} identify inherent bias amplification as a result of adversarial training and propose a framework to mitigate these biases. Our goals are different -- while these aim to reduce test time performance bias across groups, we analyze influence of minority samples on majority group performance.

\noindent\textbf{Learning from multiple domains:}
Domain adaptation literature explores learning from multiple sources~\cite{redko2020survey}. It could therefore be one potential way to analyze our problem of training on combinations of majority and minority data. In our setting, data arising from distinct domains is seen as being drawn from different distributions with a domain gap~\cite{ben2007analysis}. Between these domains,~\cite{ben2010theory} establishes error bounds for learning from combinations of domains. However, these error estimates and bounds do not take into account the notion of majority and minority groups; therefore, describing the MIME effect is outside their scope.  

\noindent\textbf{Dataset diversity:}
An important push towards fairness is through analysis of dataset composition. Several works indicate the importance of diverse datasets~\cite{gebru2018datasheets,jo2020lessons}. Ryu \etal~\cite{ryu2017inclusivefacenet} note that class imbalance in the training set leads to performance reduction. Wang \etal~\cite{wang2019balanced} highlight that perfectly balanced datasets may still not lead to balanced performance. For designing medical devices,~\cite{kadambi2021achieving} emphasizes the importance of diverse datasets. Through experiments on X-ray datasets,~\cite{larrazabal2020gender} observe that imbalanced training sets adversely affect performance on the disadvantaged group. They also observe that an unbiased training set shows the best overall accuracy. However, their inferences are related empirical observations on a few medical tasks and datasets. From an application perspective, the task of remote photoplethysmography enables analysis of the bias problem. Prior work notes that camera-based heart rate estimation exhibits skin tone bias~\cite{nowara2020meta}, and~\cite{ba2021overcoming,Wang_2022_CVPR} propose synthetic augmentations to mitigate this. Additionally,~\cite{chari2020diverse,vilesov2022blending} establish that camera based heart rate estimation is fundamentally biased against dark skin tone subjects, establishing a notion of task complexity. 
While all these works recognize that data composition affects bias, none to our knowledge describe the effect of varying minority group proportions on majority group accuracy.

\section{Statistical Origins of the MIME Effect}
\label{sec:math}

\noindent For more concise exposition, we make assumptions in the main paper derivation and defer extended generality to the supplement. Assumptions include:
\begin{itemize}
    \item Assumption 1: one-dimensional data samples and binary labels, $x \in \mathbb{R}$, $y \in \{1, 2\}$. This is relevant to modern classification problems since the final classification decision is based on a one dimensional projection of the feature representation of the sample with respect to the learnt hyperplane (discussed in Figure~\ref{fig:theory_scope}, Section~\ref{sec:verifying}). Additionally, existence proof of MIME holds for more general vectorized notation, as discussed in the supplement. 
    \item Assumption 2: the binary classifier used is a perceptron: this assumption relates to real neural networks since the last layer is perceptron-like~\cite{mohri2013perceptron}.
\end{itemize}
We now introduce some key definitions that follow from these assumptions.
\newline 
\noindent \textbf{Definition 1:} (Task complexity): {\em For binary classification we define task complexity for a group of data $\theta$ as a continuous variable in $[0,1]$, such that,
    \begin{equation}
        \theta = \underset{h \in H}{\arg\min}\,\,\epsilon(h),
    \end{equation}
    where $\epsilon(h)$ is the classification error for hypothesis $h$ (the classifier), $H$ is the space of feasible hypotheses. It is noted later that this is empirically equivalent to distributional overlap. This definition is not new. Hard-sample mining~\cite{dong2017class} establishes the of use performance measures as an indicator of difficulty.}
\newline
\noindent \textbf{Definition 2:} (Majority Group): {\em Group class (i.e. group label $g = \textrm{major}$) on which the task performs better. Quantified by training a network only with majority group data and evaluating test performance: $\theta^{\textrm{major}} = \underset{h \in H}{\arg\min}\,\,\epsilon^{\textrm{major}}(h)$.} 
\newline
\noindent \textbf{Definition 3:} (Minority Group): {\em Group class (i.e. group label $g = \textrm{minor}$) on which the task performs worse. Quantified by training a network only with minority class data and evaluating test performance: $\theta^ {\textrm{minor}} = \underset{h \in H}{\arg\min}\,\,\epsilon^{\textrm{minor}}(h)$.} 
\newline
\noindent \textbf{Definition 4:} (Minority Training Ratio ($\beta$)): {\em Ratio of minority to majority samples in the data under consideration (training set, in the context of this paper).}  
\newline    
\noindent \textbf{Definition 5:} (MIME Domain Gap): {\em Measure of how classification differs for minorities and majorities. Quantified as a difference between ideal hyperplanes. Note that this definition for domain gap could be different from other definitions. In this work, domain gap should be taken to mean MIME domain gap.}  
\newline 
\noindent Empirical observations on cutting-edge machine learning tasks demonstrate the real-world applicability of the assumptions above. We now discuss three key results. For ease of understanding, we make two simplifying assumptions for Key Results 1 and 2: (i) simplified distributions that follow a symmetric Gaussian Mixture Model, and (ii) equally likely class labels, i.e. $Pr(y=1)=Pr(y=2)$. These assumptions are relaxed in Key Result 3.

\subsection*{Key Result 1: A minority sample can be more valuable for majority classifiers than another majority sample}

\noindent Our first key result shows that it can benefit performance on the majority group more if one adds minority data (instead of majority data). Consider a binary classification setting with data samples $x \in \mathbb{R}$ and labels  $y \in \{1,2\}$. Samples from the two classes are drawn from distributions with distinct means:
\begin{equation}
\begin{split}
    x &\lvert y = 1 \sim p_1 (x \lvert \mu_1, \sigma_1) \\
    x &\lvert y = 2 \sim p_2 (x \lvert \mu_2, \sigma_2).
\end{split}
\label{eq:twodist}
\end{equation}

\begin{figure*}[t]
    \centering
    \includegraphics[width=0.65\textwidth]{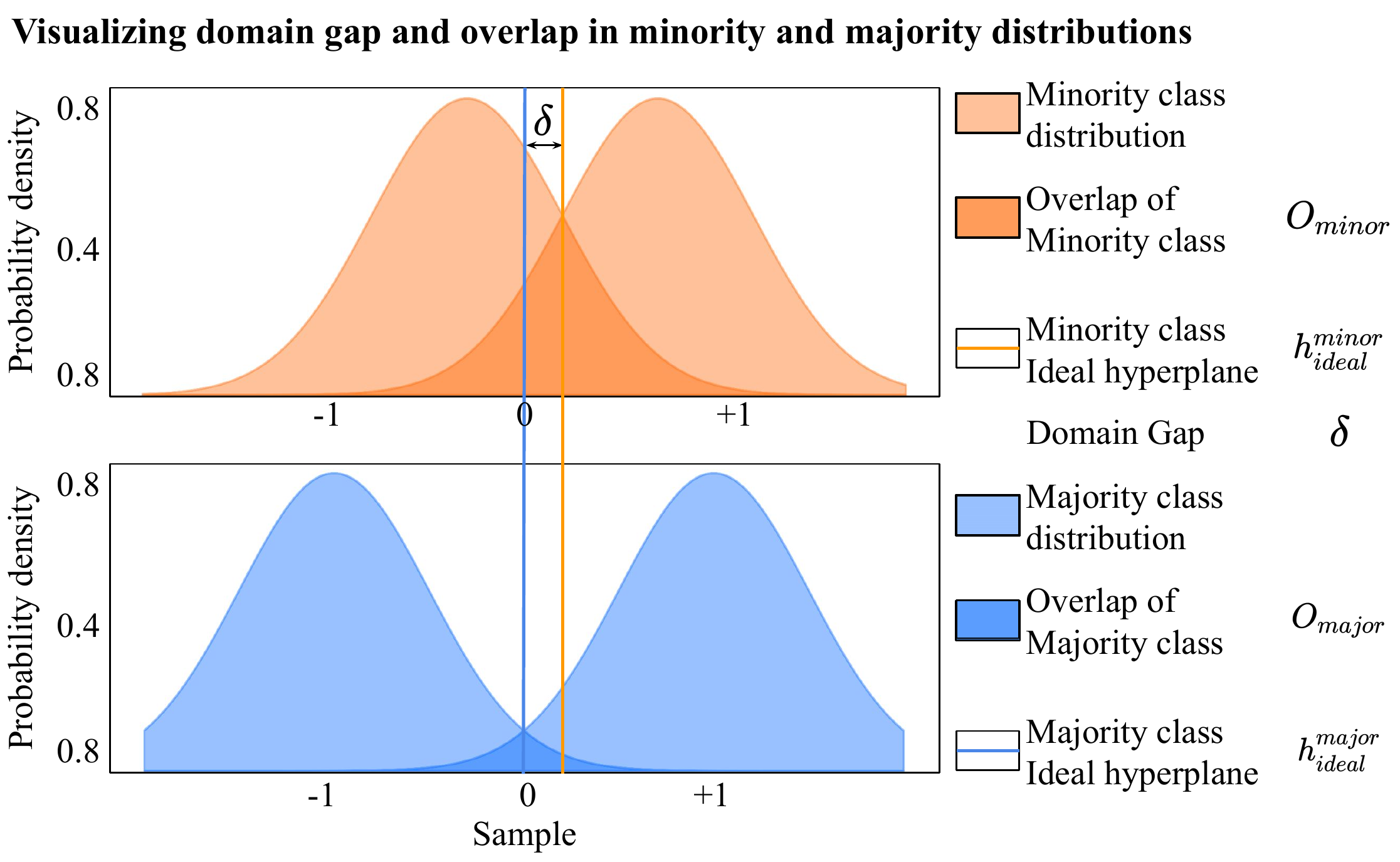}
    \caption{\textbf{Visualizating of Gaussian Mixture Model parameters.} We plot GMMs with different task complexities. The domain gap $\delta$ is visualized as the difference in the ideal threshold locations. The overlap/task complexity metric can be visually seen.}
    \label{fig:theory}
\end{figure*}

\noindent Maximum likelihood (ML) can be used to estimate the label as
\begin{equation}
    \widehat{y} = \underset{y}{\arg \max} \,\, \mathcal{L}(x \lvert y). 
\end{equation}
An ideal hyperplane for ML $\mathcal{H}_{\textrm{ideal}}$ is a set of data samples such that:
\begin{equation}
    \mathcal{H}_{\textrm{ideal}} = \big \{x \,\, \big \lvert \,\, \mathcal{L}(x \lvert y = 1) = \mathcal{L}(x \lvert y = 2) \big \}.
\end{equation}
We consider the hyperplane's geometry to be linear in this one dimensional setting. Therefore the hyperplane can be represented as a normal vector: $\mathbf{h}_\textrm{ideal}$. The normalized hyperplane is represented by a two dimensional vector, $\mathbf{h}=[1\,\,b]^T$. Here, $b$ is the offset/bias. In general, a hyperplane $\mathbf{h}$ may not be ideal. The accuracy of a hyperplane is based on a performance measure $\mathcal{P} \big \{\mathbf{h} \big \}$, where the operator $\mathcal{P}$ takes as input the hyperplane and outputs the closeness to the ideal hyperplane $\mathbf{h}_\textrm{ideal}$. A goal of a learning based classifier is to obtain:
 \begin{equation}
     \widehat{\mathbf{h}} = \underset{\mathbf{h}}{\arg \min} \,\,\mathcal{P} \big \{\mathbf{h} \big \}=\underset{\mathbf{h}}{\arg \min} \,\,\lVert \mathbf{h}-\mathbf{h}_{\textrm{ideal}} \lVert,
\end{equation}
\noindent where $\widehat{\mathbf{h}}$ is the best learnt estimate of the ideal hyperplane. The ideal hyperplane is the global minimizer of this objective. Now, assume we are provided a finite training set of labelled data $\mathcal{D}_{K-1} = \left \{ \left( x_i, y_i \right) \right\}^{K-1}_{i=1}$. Let the estimated hyperplane be $\mathbf{h}_{K-1}$, denoting that $K-1$ samples have been used to learn the hyperplane. If one additional data sample is made available, then the learnt hyperplane would be $\mathbf{h}_{K}$. From Equation~\ref{eq:twodist}, the $k$-th sample is drawn from one of two distributions: 
\begin{equation}
\begin{split}
    x_k &\lvert y = 1 \sim p_1 (x \lvert \mu_1, \sigma_1) \\
    x_k &\lvert y = 2 \sim p_2 (x \lvert \mu_2, \sigma_2).
\end{split}
\end{equation}
We now introduce the notion of majority and minority sampling.

\noindent\textbf{Introducing Majority/Minority Distributions:} Suppose that the $k$-th data sample could be drawn for the same classification task from a minority or majority group. Let $g \in \{\textrm{major},\textrm{minor} \}$ denote the group label (for the group class). Equation~\ref{eq:twodist} can now be conditioned on the group label, such that there are four possible distributions from which the $k$-th sample can be drawn:
\begin{equation}
\begin{split}
     &\left.
    \begin{array}{ll}
        x_k \lvert g=\textrm{major},y=1 \sim p^{\textrm{major}}_1 (x \lvert \mu^{\textrm{major}}_1, \sigma^{\textrm{major}}_1) \\
        x_k \lvert g=\textrm{major},y=2 \sim p^{\textrm{major}}_2 (x \lvert \mu^{\textrm{major}}_2, \sigma^{\textrm{major}}_2)
    \end{array}
    \right \} \begin{tabular}[c]{@{}c@{}}Majority\\ group\end{tabular}\\
    \\
    &\left.
    \begin{array}{ll}
        x_k \lvert g=\textrm{minor},y=1 \sim p^{\textrm{minor}}_1 (x \lvert \mu^{\textrm{minor}}_1, \sigma^{\textrm{minor}}_1) \\
      x_k \lvert g=\textrm{minor},y=2 \sim p^{\textrm{minor}}_2 (x \lvert \mu^{\textrm{minor}}_2, \sigma^{\textrm{minor}}_2) 
    \end{array}
    \right \} \begin{tabular}[c]{@{}c@{}}Minority\\ group\end{tabular}\\
\end{split}
\end{equation}

\noindent \textbf{Overlap:} Let the ideal decision hyperplane be located at $x=d_{\textrm{ideal}}$. Then, given equal likelihood of the two labels for $y$, the overlap for the majority group is defined as the probability of erroneous sample classification:
\begin{equation}
    O_{\textrm{major}}=0.5\int^{d_{\textrm{ideal}}}_{x=-\infty}p^{\textrm{major}}_2(x)dx+0.5\int^{\infty}_{x=d_{\textrm{ideal}}}p^{\textrm{major}}_1(x)dx.
\end{equation}
The same definition holds true for the minority class as well. Therefore, by definition, $O_{\textrm{major}}<O_{\textrm{minor}}$. The task complexities $\theta^{\textrm{major}}$ and $\theta^{\textrm{minor}}$ are empirical estimates of the respective overlaps.
Hereafter, we assume that all four marginal distributions are Gaussian and symmetric (this is relaxed later for Key Result 3). Figure~\ref{fig:theory} visually highlights relevant parameters. $O_{\textrm{minor}}>O_{\textrm{major}}$ occurs through the interplay of component means and variances.

\noindent The expectation over the class label yields majority and minority sampling:
\begin{equation}
    \begin{split}
            x_k^{\textrm{major}} \triangleq x_k &\lvert g=\textrm{major}\sim \mathbb{E}_y \big[x_k \lvert g=\textrm{major},y \big]\\
    x_k^{\textrm{minor}} \triangleq x_k &\lvert g=\textrm{minor}\sim \mathbb{E}_y \big[ x_k \lvert g=\textrm{minor},y \big],\\
    \end{split}
\end{equation}
where we have defined $x_k^{\textrm{major}}$ or $x_k^{\textrm{minor}}$ as having the $k$-th sample come from the majority or minority distributions. 

\noindent Armed with an expression for the $k$-th sample, we can consider a scope similar to active/online learning~\cite{ertekin2007learning,huang2010active,settles2009active,kremer2014active,dasgupta2007general,beygelzimer2010agnostic,dasgupta2011two,balcan2007margin}. Suppose a dataset of $K-1$ samples has been collected on majority samples, such that there exists a dataset stack $\mathcal{D}^{\textrm{major}}_{K-1} = \left \{ ( x_i^{\textrm{major}}, y^{\textrm{major}}_i ) \right\}^{K-1}_{i=1}$. A hyperplane $\mathbf{h}_{K-1}$ is learnt on this dataset and can be improved by expanding the dataset size. Consider pushing sample index $K$, denoted as $x_K$ onto the stack. Now we have a choice of pushing $x_K^{\textrm{major}}$ or $x_K^{\textrm{minor}}$, to create one of two datasets:
\begin{equation}
\begin{split}
    \mathcal{D}_K^{+} &= \{ \mathcal{D}_{K-1}^{\textrm{major}}, x_K^{\textrm{major}} \} \\ 
      \mathcal{D}_K^{-} &= \{ \mathcal{D}_{K-1}^{\textrm{major}}, x_K^{\textrm{minor}} \}, 
\end{split}
\end{equation}
where $\mathcal{D}_K^-$ represents the interesting case where we choose to push a minority sample onto a dataset with all majority samples (e.g. adding a dark skinned sample to a light skinned dataset). Denote $\mathbf{h}^+_K$ and $\mathbf{h}^-_K$ as hyperplanes learnt on $\mathcal{D}_K^+$ and $\mathcal{D}_K^-$. We now arrive at the following result.

\noindent \textbf{Theorem 1:} {\em Let $\mathcal{P}^{\textrm{major}} \big\{ \cdot \big\}$ be the performance of a hyperplane on the majority group. Let $\Delta=\mathcal{P}^{\textrm{major}} \big\{ \mathbf{h}_{K-1}\}$. Assume that the minority group distribution has an overlap $O_{\textrm{minor}}$ while the majority group has an overlap $O_{\textrm{major}}<O_{\textrm{minor}}$. Both have the same ideal hyperplane $\mathbf{h}_{\textrm{ideal}}$.
 Under the definitions of $\mathbf{h}^-_K$  and $\mathbf{h}^+_K$ as above, assuming $\Delta$ is sufficiently small and the group class distribution variances are not very large, 
 
 \begin{equation}
   \underset{x_K^{\textrm{minor}}}{\mathbb{E}} \mathcal{P}^{\textrm{major}} \big\{ \mathbf{h}^-_K \big\} <\underset{x_K^{\textrm{major}}}{\mathbb{E}} \mathcal{P}^{\textrm{major}} \big\{ \mathbf{h}^+_K \big\},
\end{equation}

\noindent stating that, perhaps surprisingly, expected performance for majorities improves more by pushing a minority sample on the stack, rather than a majority sample.} 
\newline 
\noindent \textbf{Proof (Sketch):} 
{\em
A sketch is provided, please see the supplement for the full proof. The general idea is to show that samples closer to $\mathbf{h}_{\textrm{ideal}}$ are more beneficial, and minority distributions may sample these with higher likelihood. Without loss of generality, we assume that $\mathbf{h}_{K-1}$ is located, non-ideally, closer to the task class $y=2$ (arbitrarily called the positive class) than $\mathbf{h}_{\textrm{ideal}}$. For our perceptron update rule, the improvement in the estimated hyperplane due to $x_K$ is proportional to the difference between the false negative rate (FNR) and the false positive rate (FPR) for $\mathbf{h}_{K-1}$, with respect to the distribution of $x_K$. For sufficiently small $\Delta$, $FNR-FPR$ can be approximated in terms of the likelihood $l$ that $x_K$ is on the ideal hyperplane. The likelihood $l$ is directly proportional to $FPR-FNR$. Under the assumptions of the theorem, a direct relation is established between the overlap and $l$ for each of the group classes. Then, it is shown that an additional minority sample, with overlap $O_{\textrm{minor}}>O_{\textrm{major}}$ leads to greater expected gains as compared to an additional majority sample, concluding the proof. $\blacksquare$
}

\subsection*{Key Result 2: MIME holds under domain gap}

\noindent In the previous key result we described the MIME effect in a restrictive setting where a minority and majority group have the same target hyperplane. However, it is rarely the case that minorities and majorities have the same decision boundary. We now consider the case with non-zero domain gap, to show that MIME holds on a more realistic setting. Domain gap can be quantified in terms of ideal decision hyperplanes. If $\mathbf{h^{\textrm{major}}_{\textrm{ideal}}}$ and $\mathbf{h^{\textrm{minor}}_{\textrm{ideal}}}$ denote ideal hyperplanes for the majority and minority groups respectively, then domain gap 
$\delta = \lVert\mathbf{h^{\textrm{major}}_{\textrm{ideal}}}-\mathbf{h^{\textrm{minor}}_{\textrm{ideal}}}\lVert$.

\noindent A visual illustration of domain gap is provided in Figure~\ref{fig:theory}. Next, we define relative hyperplane locations in terms of halfspaces (since all hyperplanes in the one dimensional setting are parallel). We say two hyperplanes $\mathbf{h}_1$ and $\mathbf{h}_2$ lie in the same halfspace of a reference hyperplane $\mathbf{h}_0$ if their respective offsets/biases satisfy the condition $(b_1-b_0)(b_2-b_0)>0$. For occupancy in different halfspaces, the condition is $(b_1-b_0)(b_2-b_0)<0$.
We now enter into the second key result. 
\newline 

 \noindent \textbf{Theorem 2:} {\em Let $\delta\neq 0$ be the domain gap between the majority and minority groups. 
 Assume that the minority group distribution has an ideal hyperplane $\mathbf{h}^{\textrm{minor}}_{\textrm{ideal}}$; while the majority group has an ideal hyperplane $\mathbf{h}^{\textrm{major}}_{\textrm{ideal}}$. 
 Then, if $\delta<\Delta$, $\delta+\Delta$ is small enough, and the group class distribution variances are not very large, it can be shown that if either of the following two cases:
 \begin{enumerate}
     \item $\mathbf{h}_{K-1}$ and $\mathbf{h}^{\textrm{minor}}_{ideal}$ lie in different halfspaces of $\mathbf{h}^{\textrm{major}}_{ideal}$,\\
     or
     \item $\mathbf{h}_{K-1}$ and $\mathbf{h}^{\textrm{minor}}_{ideal}$ lie in the same halfspace of $\mathbf{h}^{\textrm{major}}_{ideal}$, and if
     \begin{equation}
         \frac{O_{\textrm{major}}}{O_{\textrm{minor}}}<(1-\frac{\delta}{\Delta})f,
     \end{equation}
 \end{enumerate}
\noindent are true, then:
\begin{equation}
   \underset{x_K^{\textrm{minor}}}{\mathbb{E}} \mathcal{P}^{\textrm{major}} \big\{ \mathbf{h}^-_K \big\} <\underset{x_K^{\textrm{major}}}{\mathbb{E}} \mathcal{P}^{\textrm{major}} \big\{ \mathbf{h}^+_K \big\},
\end{equation} 
where $f$ is a non-negative constant that depends on the majority and minority means and standard deviations for all the individual GMM components.}
\newline  
\noindent \textbf{Proof (Sketch):} 
{\em
A sketch is provided, please see the supplement for the full proof. We prove independently for both cases. 
\begin{enumerate}
    \item When $\mathbf{h}_{K-1}$ and $\mathbf{h}^{\textrm{minor}}_{ideal}$ lie in different halfspaces of $\mathbf{h}^{\textrm{major}}_{ideal}$, it can be shown that the expected improvement in the hyperplane is higher for the minority group as compared to the majority group, using a similar argument as in Theorem 1. This proves the theorem for Case 1.
    
    \item When $\mathbf{h}_{K-1}$ and $\mathbf{h}^{\textrm{minor}}_{ideal}$ lie in the same halfspace of $\mathbf{h}^{\textrm{major}}_{ideal}$, and assuming that $\mathbf{h}_{K-1}$ is located closer to the positive class, we approximate the $FNR-FPR$ value as function of $\delta$, $\Delta$ and the likelihood $l$ as defined for Theorem 1. Then, through algebraic manipulation, constraints can be established in terms of the two likelihoods $l_{\textrm{minor}}$ and $l_{\textrm{major}}$. Under the assumptions of the theorem, a relation can be established between the ratios $\frac{l_{\textrm{minor}}}{l_{\textrm{major}}}$ and $\frac{O_{\textrm{minor}}}{O_{\textrm{major}}}$. This proves the theorem for Case 2, and concludes the proof. $\blacksquare$
\end{enumerate}
}

\subsection*{Key Result 3: MIME holds for general distributions}
We now relax the symmetric Gaussian and equally likely labels requirements to arrive at a general condition for MIME existence. Let $p^{\textrm{major}}_1$ and $p^{\textrm{major}}_2$ be general distributions describing the majority group $y=1$ and $y=2$ classes. Additionally, $Pr(y=1)\neq Pr(y=2)$. Minority group distributions are described similarly. We define the signed tail weight for the majority group as follows:
\begin{equation}
    T^{\textrm{major}}(x_d)=  \pi^{\textrm{major}}\int^{x_d}_{x=-\infty}p^{\textrm{major}}_2(x)dx-(1-\pi^{\textrm{major}})\int^{\infty}_{x=x_d}p^{\textrm{major}}_1(x)dx,
\end{equation}
where $\pi^{\textrm{major}}=Pr(x=2)$ for the majority group. $T^{\textrm{minor}}(\cdot)$ is similarly defined.
This leads us to our third key result.
\newline 

 \noindent \textbf{Theorem 3:} {\em Consider majority and minority groups, with general sample distributions and unequal prior label distributions. 
 If,
 \begin{multline}
     \textrm{min}\left\{ T^{\textrm{minor}}(d_{\textrm{ideal}}+\Delta), -T^{\textrm{minor}}(d_{\textrm{ideal}}-\Delta)\right\}>\\
     \textrm{max}\left\{ T^{\textrm{major}}(d_{\textrm{ideal}}+\Delta), -T^{\textrm{major}}(d_{\textrm{ideal}}-\Delta)\right\},
     \label{eq:th3}
 \end{multline}
then $\underset{x_K^{\textrm{minor}}}{\mathbb{E}} \mathcal{P}^{\textrm{major}} \big\{ \mathbf{h}^-_K \big\} <\underset{x_K^{\textrm{major}}}{\mathbb{E}} \mathcal{P}^{\textrm{major}} \big\{ \mathbf{h}^+_K \big\}$.
}
\newline  
\noindent \textbf{Proof (Sketch):} 
{\em
A sketch is provided, please see the supplement for the full proof. The perceptron algorithm update rule is proportional to $FNR-FPR$ (if $\mathbf{h}_{K-1}$ is located closer to the positive class) or the $FPR-FNR$ (if $\mathbf{h}_{K-1}$ is located closer to the negative class). The MIME effect exists in the scenario where the worst case update for the minority group is better than the best case update for the majority group (described in Equation~\ref{eq:th3}). This proves the theorem. $\blacksquare$
}

\noindent Generalizations of Theorem 3 to include domain gap are discussed in the supplement, for brevity. Theorems 1 and 2 are special cases of the general Theorem 3, describing MIME existence for specific group distributions.

\section{Verifying MIME Theory on Real Tasks}
\label{sec:verifying}
\begin{figure*}[t]
    \centering
    \includegraphics[width=\textwidth]{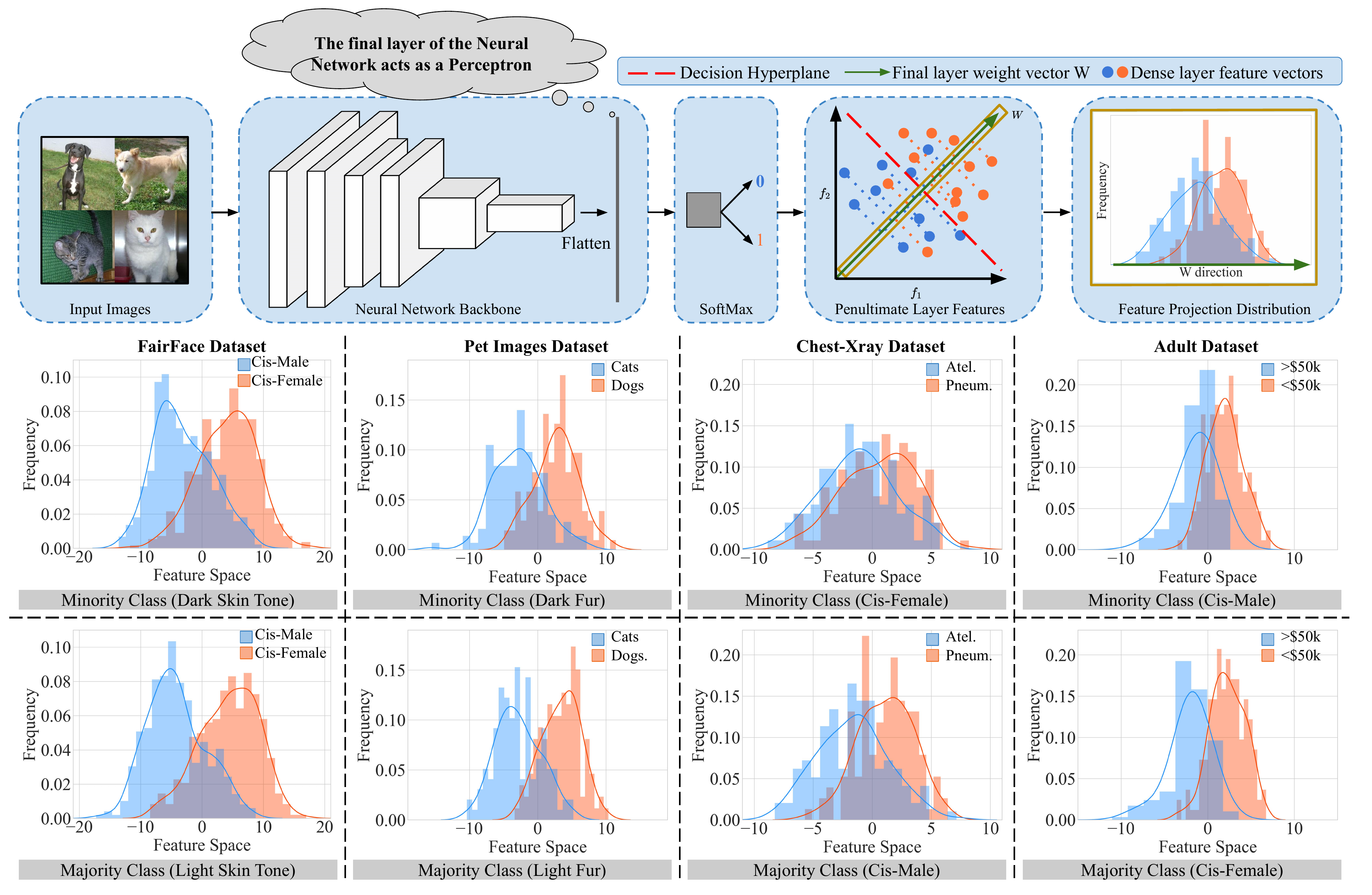}
    \caption{\textbf{The use of Gaussian mixtures to represent minority and majority distributions is consistent with behaviors in modern neural networks, on real-world datasets.} (top row) The last layer of common neural architectures is a linear classifier on features. Histograms of the penultimate layer projections are generated for models with $\beta=0.5$. (middle row) Minority histograms: note the greater difficulty due to less separation of data. (bottom row) Majority histograms: note smaller overlap and easier classification. Figure can be parsed on a per-dataset basis. Within each column, the reader can compare the domain gap and overlap in the two histograms.}
    \label{fig:f_feature_analysis}
\end{figure*}

\noindent In the previous section, we provide existence conditions for the MIME phenomenon for general sample distributions. However, experimental validation of the phenomenon requires quantification in terms of measurable quantities such as overlap. Theorem 2 provides us these resources. Here, we verify that the assumptions in Theorem 2 are validated by experiments on real tasks. 

\subsection{Verifying Assumptions}

\begin{table}[t]
    \caption{\textbf{Experimental measures of overlap and domain gap are consistent with the theory in Section~\ref{sec:math}}. Note that the majority group consistently has lower overlap. Domain gaps are found to be small. DS-1 is FairFace, DS-2 is Pet Images, DS-4 is Chest-Xray14 and DS-5 is Adult. DS-6 is the high domain gap gender classification experiment. DS-3 is excluded here since it deals with a 9 class classification problem.}
    \footnotesize
   
  \begin{center}
    \setlength{\tabcolsep}{0.01\columnwidth}
    
\resizebox{0.7\columnwidth}{!}{
  \begin{tabular}{lcccccc} 
    \toprule
    \multirow{2}{*}{\begin{tabular}{@{}c@{}}{Dataset}\\
    {(Task)}\end{tabular}} & \multirow{2}{*}{\begin{tabular}{@{}c@{}}{DS-1~\cite{karkkainen2021fairface}}\\
    {(Gender)}\end{tabular}} &
    \multirow{2}{*}{\begin{tabular}{@{}c@{}}{DS-2~\cite{golle2008machine}}\\
    {(Species)}\end{tabular}} &
    \multirow{2}{*}{\begin{tabular}{@{}c@{}}{DS-4~\cite{rajpurkar2017chexnet}}\\
    {(Diagnosis)}\end{tabular}} &
    \multirow{2}{*}{\begin{tabular}{@{}c@{}}{DS-5~\cite{blake1998uci}}\\
    {(Income)}\end{tabular}}&
    \multirow{2}{*}{\begin{tabular}{@{}c@{}}{DS-6}~\cite{zhang2017age,yao2020estimation}\\
    {(Gender)}\end{tabular}}\\
    &&&&&&\\
    \midrule
    Major. overlap & 0.186 & 0.163 & 0.294 & 0.132&0.09\\
    Minor. overlap & 0.224 & 0.198 & 0.369 & 0.208&0.19\\
    Domain gap & 0.276 & 0.518 & 0.494 & 0.170 & 1.62\\
    \bottomrule
  \end{tabular} } 
 \end{center}
  \label{tab:emprical_analysis}
\end{table}

\noindent\textbf{Verifying Gaussianity:} Theorem 2 assumes that data $x$ is drawn from a Gaussian Mixture Model. At first glance, this quantification may appear to be unrelated to complex neural networks. However, as illustrated at the top of Figure~\ref{fig:f_feature_analysis}, a ConvNet is essentially a feature extractor that feeds a flattened layer into a simple perceptron or linear classifier. The flattened layer can be orthogonally projected onto the decision boundary to generate, in analogy, an $x$ used for linear classification (Figure~\ref{fig:theory_scope}, fixed-backbone configuration). We use this as a first approximation to the end-to-end configuration used in our experiments.

\noindent Plotting empirical histograms of these flattened layers (Figure~\ref{fig:f_feature_analysis}) shows Gaussian-like distribution. This is consistent with the Law of Large Numbers -- linear combination of several random variables follows an approximate Gaussian distribution. Hence, Theorem 2 is approximately related in this setting. Details about implementation and comparison to Gaussians are deferred to the supplement. 

\noindent\textbf{Verifying minority/majority definitions:} The MIME proof linked minority and majority definitions to distributional overlap and domain gap. Given the histogram embeddings from above, it is seen that minority groups on all four vision tasks have greater overlap. There also exists a domain gap between majority and minority but this is small compared to distribution spread (except for the high domain gap experiment). This establishes applicability of small domain gap requirements.  Quantification is provided in Table~\ref{tab:emprical_analysis}. Code is in the supplement. 
\subsection{MIME Effect Across Six, Real Datasets}
\label{ss:fiveresults}

\begin{figure*}[t]
    \centering
    \includegraphics[width=\textwidth]{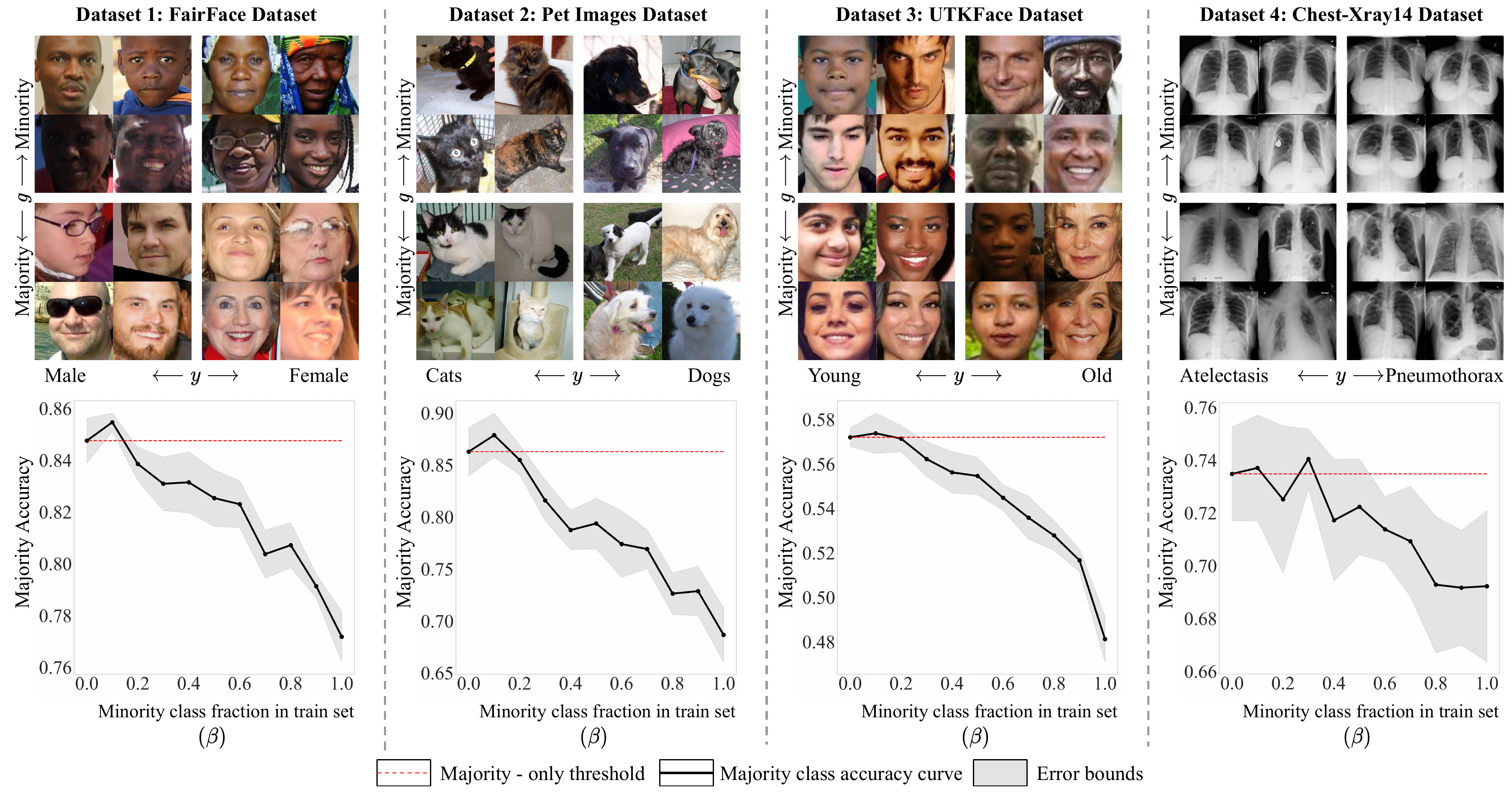}
    \caption{\textbf{When domain gap is small, the MIME effect holds.} On four vision datasets, majority performance is maximized with some inclusion of minorities. All experiments are run for several trials and realizations (described in Section~\ref{ss:fiveresults}).}
    \label{fig:visual-datasets}
\end{figure*}

\noindent\textbf{Implementation:} Six multi-attribute datasets are used to assess the MIME effect (five are in computer vision). For a particular \textit{experiment}, we identify a task category to evaluate accuracy over (e.g. gender), and a group category (e.g. race). The best test accuracy on the majority group across all epochs is recorded as our accuracy measure. Each experiment is run for a fixed number of \textit{minority training ratios} ($\beta$). For each minority training ratio, the total number of training samples remains constant. That is, the minority samples replace the majority samples, instead of being appended to the training set. Each experiment is also run for a finite number of \textit{trials}. Different trials have different random train and test sets (except for the FairFace dataset~\cite{karkkainen2021fairface} where we use the provided test split). Averaging is done across trials. Note that minority samples to be added are randomly chosen -- the MIME effect is not specific to particular samples. For the vision datasets, we use a ResNet-34 architecture~\cite{he2016deep}, with the output layer appropriately modified. For the non-visual dataset, a fully connected network is used. Average accuracy and trend error, across trials are used to evaluate performance. Specific implementation details are in the supplement. 

\noindent\textbf{MIME effect on gender  classification:} The FairFace dataset~\cite{karkkainen2021fairface} is used to perform gender classification ($y=1$ is male, $y=2$ is female). The majority and minority groups $g = \{\textrm{major}, \textrm{minor} \}$ are light and dark skin, respectively. Results are averaged over five trials. Figure~\ref{fig:visual-datasets} describes qualitative accuracy. The accuracy trends indicate that adding 10\% of minority samples to the training set leads to approximately a 1.5\% gain in majority group (light skin) test accuracy.

\noindent\textbf{MIME effect on animal species identification:}
 We manually annotate light and dark cats and dogs from the Pets dataset~\cite{golle2008machine}.
 We classify between cats ($y=1$) and dogs ($y=2$). The majority and minority groups are light and dark fur color respectively. Figure~\ref{fig:visual-datasets} shows qualitative results. Over five trials, we see a majority group accuracy gain of about 2\%, with a peak at $\beta=$10\%.

\noindent\textbf{MIME effect on age classification:}
We use a second human faces dataset, the UTKFace dataset~\cite{zhang2017age}, for the age classification task (9 classes of age-intervals). We pre-process the UTKFace age labels into class bins to match the FairFace dataset format. The majority and minority groups are male and female respectively. The proportion of task class labels is kept the same across group classes. Results are averaged over five trials. Figure~\ref{fig:visual-datasets} shows trends. We observe a smaller average improvement for the 10\% minority training ratio. However, since these are average trends, this indicates consistent gain. Results on this dataset also empirically highlight the existence of the MIME effect beyond two class settings.

\noindent\textbf{MIME effect on X-ray diagnosis Classification:}
We use the NIH Chest-Xray14 dataset~\cite{rajpurkar2017chexnet} to analyze trends on a medical imaging task. We perform binary classification of scans belonging to `Atelectasis' ($y=1$) and `Pneumothorax' ($y=2$) categories. The male and female genders are the majority and minority groups respectively. Results are averaged over seven trials (due to noisier trends). From Figure~\ref{fig:visual-datasets}, we observe noisy trends - specifically we see a performance drop for $\beta=0.2$, prior to an overall gain for $\beta=0.3$. The error bounds also have considerably more noise. However, confidence in the peak and the MIME effect, as seen from the average trends and the error bounds, remains high.

\begin{figure*}[t]
    \centering
    \includegraphics[width=\textwidth]{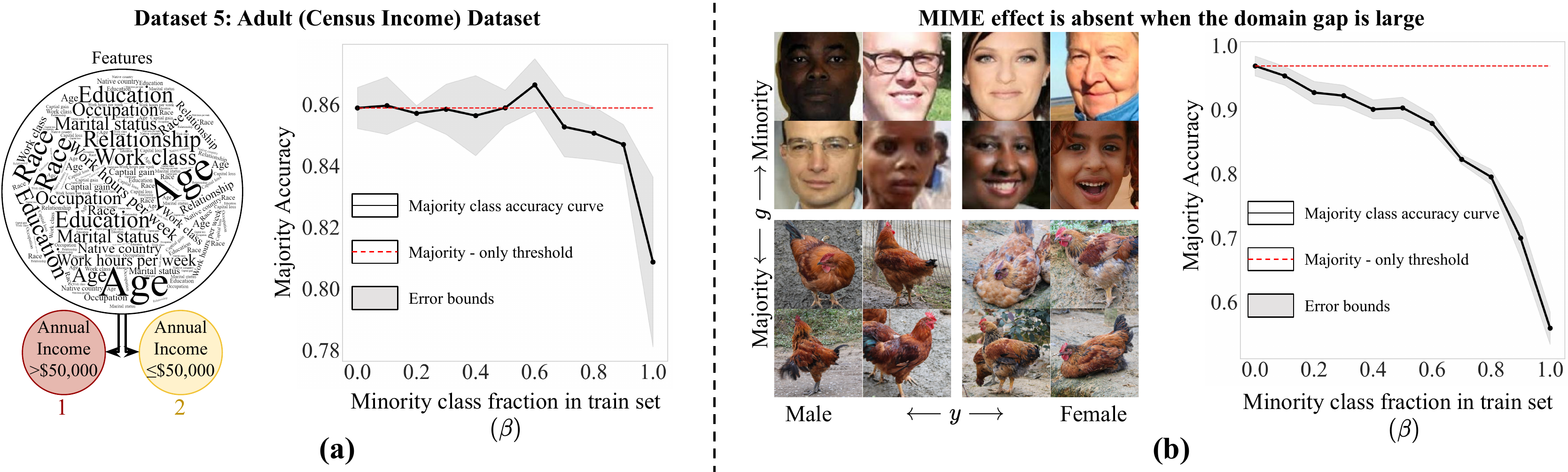}
    \caption{\textbf{MIME effect is observed in non-vision datasets, and is absent in the case of large domain gap.} (a) The Adult Dataset~\cite{blake1998uci} uses Census data to predict an income label. (b) On dataset six, gender classification is rescoped to occur in a high domain gap setting. Majority group is chickens~\cite{yao2020estimation} and minority group is humans~\cite{zhang2017age}.}
    \label{fig:data2}
\end{figure*}

\noindent\textbf{MIME effect on income classification:} For validation in a non-vision setting, we use the Adult (Census Income) dataset~\cite{blake1998uci}. The data consists of census information with annual income labels (income less than or equal to \$50,000 is $y=1$, income greater than \$50,000 is $y=2$). The majority and minority groups are female and male genders respectively. Results are averaged over five trials. Figure~\ref{fig:data2}(a) highlights a prominent accuracy gain for $\beta=0.6$. 

\noindent\textbf{MIME effect and domain gap:} Theorem 2 (Section~\ref{sec:math}) suggests that large domain gap settings will not show the MIME effect. We set up an experiment to verify this (Figure~\ref{fig:data2}(b)). Gender classification among chickens (majority group) and humans (minority group) has a high domain gap due to minimal common context (validated by the domain gap estimates, Table~\ref{tab:emprical_analysis}). With increasing $\beta$, the majority accuracy decreases. This (and Figure~\ref{fig:f_feature_analysis}, Table~\ref{tab:emprical_analysis} that show low domain gap for other datasets) validates Theorem 2. Note that while this result may not be unexpected, it further validates our proposed theory.

\section{Discussion}
\label{sec:discussion}

\noindent\textbf{Secondary validation and analysis:} Table~\ref{tab:additional_validation} supplies additional metrics to analyze MIME. Across datasets, almost all trials show existence, with every dataset showing average MIME performance gain. Some readers may view the error bars in Figures~\ref{fig:visual-datasets} and~\ref{fig:data2} as large, however they are comparable to other empirical ML works~\cite{liu2021iterative,d2021interplay}; they may appear larger due to scaling. Reasons for error bars include variations in train-test data and train set size (Table~B and C, supplement). Further analysis, including interplay with debiasing methods (e.g. hard-sample mining~\cite{dong2017class}) and reconciliation with work on equal representation datasets~\cite{gebru2018datasheets,buolamwini2018gender,larrazabal2020gender,ryu2017inclusivefacenet,li2019repair,mehrabi2021survey,jo2020lessons,gong2019diversity,kadambi2021achieving} is deferred to the supplement.

\begin{table}[t]
    \caption{\textbf{Additional evaluation metrics provide further evidence of MIME existence across all datasets.} The table highlights: (i) number of trials with MIME performance gain (i.e. majority accuracy at some $\beta>0$ is greater than majority accuracy at $\beta=0$), and (ii) the mean MIME performance gain across trials (in \% points).} 
    \footnotesize
   
  \begin{center}
    \setlength{\tabcolsep}{0.01\columnwidth}
    
\resizebox{0.7\columnwidth}{!}{
  \begin{tabular}{lcccccc} 
    \toprule
    Dataset & {DS-1~\cite{karkkainen2021fairface}} &
    {DS-2~\cite{golle2008machine}} &
    {DS-3~\cite{zhang2017age}} &
    {DS-4~\cite{rajpurkar2017chexnet}} &
    {DS-5~\cite{blake1998uci}}\\
    \midrule
    \#MIME trials/Total trials & 4/5 & 4/5 & 5/5 & 6/7 & 4/5\\
    Avg. MIME perf. gain & 0.72\% & 1.84\% & 0.70\% & 1.89\% & 0.98\%  \\
    \bottomrule
  \end{tabular} } 
 \end{center}
  \label{tab:additional_validation}
\end{table}

\noindent\textbf{Optimality of inclusion ratios:} Our experiments show that there can exist an optimal amount of minority inclusion to benefit the majority group the most. This appears true across all experiments in Figures~\ref{fig:visual-datasets},~\ref{fig:data2}. However, beyond a certain amount, accuracy decreases consistently, with lowest accuracy on majority samples observed when no majorities are used in training. This optimal $\beta$ depends on individual task complexities, among other factors. Since identifying it is outside our scope (Section~\ref{sec:contributions},~\ref{sec:theory_scope}), our experiments use $10\%$ sampling resolution for $\beta$. Peaks at $\beta=10\%$ for some datasets are due to this lower resolution; optimal peak need not lie there for all datasets (e.g. X-ray~\cite{rajpurkar2017chexnet} \& Adult~\cite{blake1998uci}). Future work can identify optimal ratios through finer analysis over $\beta$.

\noindent\textbf{Limitations:} The theoretical scope is certifiable within fixed-backbone binary classification, which is narrower than all of machine learning (Figure~\ref{fig:theory_scope}). Should this theory be accepted by the community, follow-up work can generalize theoretical claims. Another limitation is the definition-compatibility of majority and minority groups. Our theory is applicable to task-advantage definitions; some scholars in the community instead define majorities and minorities by proportion. Our theory is applicable to these authors as well, albeit with a slight redefinition of terminology. Additional considerations are included in the supplement.

\noindent\textbf{Conclusion:} In conclusion, majority performance benefits from a non-zero fraction of inclusion of minority data given a sufficiently small domain gap. 

\subsubsection{Acknowledgements} We thank members of the Visual Machines Group for their feedback and support. A.K. was partially supported by an NSF CAREER award IIS-2046737 and Army Young Investigator Award. P.C. was partially supported by a Cisco PhD Fellowship.

\clearpage
%
%
\bibliographystyle{splncs04}
\bibliography{egbib} 

\begin{thebibliography}{10}
\providecommand{\url}[1]{\texttt{#1}}
\providecommand{\urlprefix}{URL }
\providecommand{\doi}[1]{https://doi.org/#1}

\bibitem{ba2021overcoming}
Ba, Y., Wang, Z., Karinca, K.D., Bozkurt, O.D., Kadambi, A.: Overcoming
  difficulty in obtaining dark-skinned subjects for remote-ppg by synthetic
  augmentation. arXiv preprint arXiv:2106.06007  (2021)

\bibitem{balakrishnan2021towards}
Balakrishnan, G., Xiong, Y., Xia, W., Perona, P.: Towards causal benchmarking
  of biasin face analysis algorithms. In: Deep Learning-Based Face Analytics,
  pp. 327--359. Springer (2021)

\bibitem{balcan2007margin}
Balcan, M.F., Broder, A., Zhang, T.: Margin based active learning. In:
  International Conference on Computational Learning Theory. pp. 35--50.
  Springer (2007)

\bibitem{bellamy2019ai}
Bellamy, R.K., Dey, K., Hind, M., Hoffman, S.C., Houde, S., Kannan, K., Lohia,
  P., Martino, J., Mehta, S., Mojsilovi{\'c}, A., et~al.: Ai fairness 360: An
  extensible toolkit for detecting and mitigating algorithmic bias. IBM Journal
  of Research and Development  \textbf{63}(4/5), ~4--1 (2019)

\bibitem{ben2010theory}
Ben-David, S., Blitzer, J., Crammer, K., Kulesza, A., Pereira, F., Vaughan,
  J.W.: A theory of learning from different domains. Machine learning
  \textbf{79}(1),  151--175 (2010)

\bibitem{ben2007analysis}
Ben-David, S., Blitzer, J., Crammer, K., Pereira, F., et~al.: Analysis of
  representations for domain adaptation. Advances in Neural Information
  Processing Systems  \textbf{19}, ~137 (2007)

\bibitem{beygelzimer2010agnostic}
Beygelzimer, A., Hsu, D.J., Langford, J., Zhang, T.: Agnostic active learning
  without constraints. Advances in Neural Information Processing Systems
  \textbf{23},  199--207 (2010)

\bibitem{bickel2009discriminative}
Bickel, S., Br{\"u}ckner, M., Scheffer, T.: Discriminative learning under
  covariate shift. Journal of Machine Learning Research  \textbf{10}(9) (2009)

\bibitem{blake1998uci}
Blake, C.L., Merz, C.J.: Uci repository of machine learning databases, 1998
  (1998)

\bibitem{bolukbasi2016man}
Bolukbasi, T., Chang, K.W., Zou, J.Y., Saligrama, V., Kalai, A.T.: Man is to
  computer programmer as woman is to homemaker? debiasing word embeddings.
  Advances in Neural Information Processing Systems  \textbf{29},  4349--4357
  (2016)

\bibitem{buolamwini2018gender}
Buolamwini, J., Gebru, T.: Gender shades: Intersectional accuracy disparities
  in commercial gender classification. In: Conference on fairness,
  accountability and transparency. pp. 77--91. PMLR (2018)

\bibitem{chari2020diverse}
Chari, P., Kabra, K., Karinca, D., Lahiri, S., Srivastava, D., Kulkarni, K.,
  Chen, T., Cannesson, M., Jalilian, L., Kadambi, A.: Diverse r-ppg:
  Camera-based heart rate estimation for diverse subject skin-tones and scenes.
  arXiv preprint arXiv:2010.12769  (2020)

\bibitem{choi2021active}
Choi, J., Elezi, I., Lee, H.J., Farabet, C., Alvarez, J.M.: Active learning for
  deep object detection via probabilistic modeling. arXiv preprint
  arXiv:2103.16130  (2021)

\bibitem{cui2019class}
Cui, Y., Jia, M., Lin, T.Y., Song, Y., Belongie, S.: Class-balanced loss based
  on effective number of samples. In: Proceedings of the IEEE/CVF Conference on
  Computer Vision and Pattern Recognition. pp. 9268--9277 (2019)

\bibitem{d2021interplay}
d'Ascoli, S., Gabri{\'e}, M., Sagun, L., Biroli, G.: On the interplay between
  data structure and loss function in classification problems. Advances in
  Neural Information Processing Systems  \textbf{34},  8506--8517 (2021)

\bibitem{dasgupta2011two}
Dasgupta, S.: Two faces of active learning. Theoretical computer science
  \textbf{412}(19),  1767--1781 (2011)

\bibitem{dasgupta2007general}
Dasgupta, S., Hsu, D.J., Monteleoni, C.: A general agnostic active learning
  algorithm. Citeseer (2007)

\bibitem{dong2017class}
Dong, Q., Gong, S., Zhu, X.: Class rectification hard mining for imbalanced
  deep learning. In: Proceedings of the IEEE/CVF International Conference on
  Computer Vision. pp. 1851--1860 (2017)

\bibitem{elkan2001foundations}
Elkan, C.: The foundations of cost-sensitive learning. In: International joint
  conference on artificial intelligence. vol.~17, pp. 973--978. Lawrence
  Erlbaum Associates Ltd (2001)

\bibitem{ertekin2007learning}
Ertekin, S., Huang, J., Bottou, L., Giles, L.: Learning on the border: active
  learning in imbalanced data classification. In: Proceedings of the sixteenth
  ACM conference on Conference on information and knowledge management. pp.
  127--136 (2007)

\bibitem{gebru2018datasheets}
Gebru, T., Morgenstern, J., Vecchione, B., Vaughan, J.W., Wallach, H.,
  Daum{\'e}~III, H., Crawford, K.: Datasheets for datasets. arXiv preprint
  arXiv:1803.09010  (2018)

\bibitem{golle2008machine}
Golle, P.: Machine learning attacks against the asirra captcha. In: Proceedings
  of the 15th ACM conference on Computer and communications security. pp.
  535--542 (2008)

\bibitem{gong2019diversity}
Gong, Z., Zhong, P., Hu, W.: Diversity in machine learning. IEEE Access
  \textbf{7},  64323--64350 (2019)

\bibitem{gwilliam2021rethinking}
Gwilliam, M., Hegde, S., Tinubu, L., Hanson, A.: Rethinking common assumptions
  to mitigate racial bias in face recognition datasets. In: Proceedings of the
  IEEE/CVF International Conference on Computer Vision. pp. 4123--4132 (2021)

\bibitem{he2016deep}
He, K., Zhang, X., Ren, S., Sun, J.: Deep residual learning for image
  recognition. In: Proceedings of the IEEE/CVF Conference on Computer Vision
  and Pattern Recognition. pp. 770--778 (2016)

\bibitem{hendricks2018women}
Hendricks, L.A., Burns, K., Saenko, K., Darrell, T., Rohrbach, A.: Women also
  snowboard: Overcoming bias in captioning models. In: Proceedings of the
  European Conference on Computer Vision (ECCV). pp. 771--787 (2018)

\bibitem{huang2021fsdr}
Huang, J., Guan, D., Xiao, A., Lu, S.: Fsdr: Frequency space domain
  randomization for domain generalization. In: Proceedings of the IEEE/CVF
  Conference on Computer Vision and Pattern Recognition. pp. 6891--6902 (2021)

\bibitem{huang2010active}
Huang, S.J., Jin, R., Zhou, Z.H.: Active learning by querying informative and
  representative examples. Advances in Neural Information Processing Systems
  \textbf{23},  892--900 (2010)

\bibitem{jo2020lessons}
Jo, E.S., Gebru, T.: Lessons from archives: Strategies for collecting
  sociocultural data in machine learning. In: Proceedings of the 2020
  Conference on Fairness, Accountability, and Transparency. pp. 306--316 (2020)

\bibitem{kadambi2021achieving}
Kadambi, A.: Achieving fairness in medical devices. Science
  \textbf{372}(6537),  30--31 (2021)

\bibitem{kang2019decoupling}
Kang, B., Xie, S., Rohrbach, M., Yan, Z., Gordo, A., Feng, J., Kalantidis, Y.:
  Decoupling representation and classifier for long-tailed recognition. arXiv
  preprint arXiv:1910.09217  (2019)

\bibitem{karkkainen2021fairface}
Karkkainen, K., Joo, J.: Fairface: Face attribute dataset for balanced race,
  gender, and age for bias measurement and mitigation. In: Proceedings of the
  IEEE/CVF Winter Conference on Applications of Computer Vision. pp. 1548--1558
  (2021)

\bibitem{kremer2014active}
Kremer, J., Steenstrup~Pedersen, K., Igel, C.: Active learning with support
  vector machines. Wiley Interdisciplinary Reviews: Data Mining and Knowledge
  Discovery  \textbf{4}(4),  313--326 (2014)

\bibitem{larrazabal2020gender}
Larrazabal, A.J., Nieto, N., Peterson, V., Milone, D.H., Ferrante, E.: Gender
  imbalance in medical imaging datasets produces biased classifiers for
  computer-aided diagnosis. Proceedings of the National Academy of Sciences
  \textbf{117}(23),  12592--12594 (2020)

\bibitem{li2019repair}
Li, Y., Vasconcelos, N.: Repair: Removing representation bias by dataset
  resampling. In: Proceedings of the IEEE/CVF Conference on Computer Vision and
  Pattern Recognition. pp. 9572--9581 (2019)

\bibitem{liu2021iterative}
Liu, T., Vietri, G., Wu, S.Z.: Iterative methods for private synthetic data:
  Unifying framework and new methods. Advances in Neural Information Processing
  Systems  \textbf{34},  690--702 (2021)

\bibitem{mehrabi2021survey}
Mehrabi, N., Morstatter, F., Saxena, N., Lerman, K., Galstyan, A.: A survey on
  bias and fairness in machine learning. ACM Computing Surveys (CSUR)
  \textbf{54}(6),  1--35 (2021)

\bibitem{mohri2013perceptron}
Mohri, M., Rostamizadeh, A.: Perceptron mistake bounds. arXiv preprint
  arXiv:1305.0208  (2013)

\bibitem{nowara2020meta}
Nowara, E.M., McDuff, D., Veeraraghavan, A.: A meta-analysis of the impact of
  skin tone and gender on non-contact photoplethysmography measurements. In:
  Proceedings of the IEEE/CVF Conference on Computer Vision and Pattern
  Recognition Workshops. pp. 284--285 (2020)

\bibitem{rajpurkar2017chexnet}
Rajpurkar, P., Irvin, J., Zhu, K., Yang, B., Mehta, H., Duan, T., Ding, D.,
  Bagul, A., Langlotz, C., Shpanskaya, K., et~al.: Chexnet: Radiologist-level
  pneumonia detection on chest x-rays with deep learning. arXiv preprint
  arXiv:1711.05225  (2017)

\bibitem{ramaswamy2021fair}
Ramaswamy, V.V., Kim, S.S., Russakovsky, O.: Fair attribute classification
  through latent space de-biasing. In: Proceedings of the IEEE/CVF Conference
  on Computer Vision and Pattern Recognition. pp. 9301--9310 (2021)

\bibitem{redko2020survey}
Redko, I., Morvant, E., Habrard, A., Sebban, M., Bennani, Y.: A survey on
  domain adaptation theory: learning bounds and theoretical guarantees. arXiv
  preprint arXiv:2004.11829  (2020)

\bibitem{ren2018learning}
Ren, M., Zeng, W., Yang, B., Urtasun, R.: Learning to reweight examples for
  robust deep learning. In: International Conference on Machine Learning. pp.
  4334--4343. PMLR (2018)

\bibitem{ryu2017inclusivefacenet}
Ryu, H.J., Adam, H., Mitchell, M.: Inclusivefacenet: Improving face attribute
  detection with race and gender diversity. arXiv preprint arXiv:1712.00193
  (2017)

\bibitem{settles2009active}
Settles, B.: Active learning literature survey  (2009)

\bibitem{tartaglione2021end}
Tartaglione, E., Barbano, C.A., Grangetto, M.: End: Entangling and
  disentangling deep representations for bias correction. In: Proceedings of
  the IEEE/CVF Conference on Computer Vision and Pattern Recognition. pp.
  13508--13517 (2021)

\bibitem{tremblay2018training}
Tremblay, J., Prakash, A., Acuna, D., Brophy, M., Jampani, V., Anil, C., To,
  T., Cameracci, E., Boochoon, S., Birchfield, S.: Training deep networks with
  synthetic data: Bridging the reality gap by domain randomization. In:
  Proceedings of the IEEE/CVF Conference on Computer Vision and Pattern
  Recognition Workshops. pp. 969--977 (2018)

\bibitem{vilesov2022blending}
Vilesov, A., Chari, P., Armouti, A., Harish, A.B., Kulkarni, K., Deoghare, A.,
  Jalilian, L., Kadambi, A.: Blending camera and 77 ghz radar sensing for
  equitable, robust plethysmography. In: ACM Trans. Graph. (SIGGRAPH) (2022)

\bibitem{wang2019balanced}
Wang, T., Zhao, J., Yatskar, M., Chang, K.W., Ordonez, V.: Balanced datasets
  are not enough: Estimating and mitigating gender bias in deep image
  representations. In: Proceedings of the IEEE/CVF International Conference on
  Computer Vision. pp. 5310--5319 (2019)

\bibitem{wang2020towards}
Wang, Z., Qinami, K., Karakozis, I.C., Genova, K., Nair, P., Hata, K.,
  Russakovsky, O.: Towards fairness in visual recognition: Effective strategies
  for bias mitigation. In: Proceedings of the IEEE/CVF Conference on Computer
  Vision and Pattern Recognition. pp. 8919--8928 (2020)

\bibitem{Wang_2022_CVPR}
Wang, Z., Ba, Y., Chari, P., Bozkurt, O.D., Brown, G., Patwa, P., Vaddi, N.,
  Jalilian, L., Kadambi, A.: Synthetic generation of face videos with
  plethysmograph physiology. In: Proceedings of the IEEE/CVF Conference on
  Computer Vision and Pattern Recognition (CVPR). pp. 20587--20596 (2022)

\bibitem{xu2021robust}
Xu, H., Liu, X., Li, Y., Jain, A., Tang, J.: To be robust or to be fair:
  Towards fairness in adversarial training. In: International Conference on
  Machine Learning. pp. 11492--11501. PMLR (2021)

\bibitem{yao2020estimation}
Yao, Y., Yu, H., Mu, J., Li, J., Pu, H.: Estimation of the gender ratio of
  chickens based on computer vision: Dataset and exploration. Entropy
  \textbf{22}(7), ~719 (2020)

\bibitem{yue2019domain}
Yue, X., Zhang, Y., Zhao, S., Sangiovanni-Vincentelli, A., Keutzer, K., Gong,
  B.: Domain randomization and pyramid consistency: Simulation-to-real
  generalization without accessing target domain data. In: Proceedings of the
  IEEE/CVF International Conference on Computer Vision. pp. 2100--2110 (2019)

\bibitem{zhang2018mitigating}
Zhang, B.H., Lemoine, B., Mitchell, M.: Mitigating unwanted biases with
  adversarial learning. In: Proceedings of the 2018 AAAI/ACM Conference on AI,
  Ethics, and Society. pp. 335--340 (2018)

\bibitem{zhang2017age}
Zhang, Z., Song, Y., Qi, H.: Age progression/regression by conditional
  adversarial autoencoder. In: Proceedings of the IEEE/CVF Conference on
  Computer Vision and Pattern Recognition. pp. 5810--5818 (2017)

\end{thebibliography}


\begin{thebibliography}{10}
\providecommand{\url}[1]{\texttt{#1}}
\providecommand{\urlprefix}{URL }
\providecommand{\doi}[1]{https://doi.org/#1}

\bibitem{blake1998uci}
Blake, C.L., Merz, C.J.: Uci repository of machine learning databases, 1998
  (1998)

\bibitem{buolamwini2018gender}
Buolamwini, J., Gebru, T.: Gender shades: Intersectional accuracy disparities
  in commercial gender classification. In: Conference on fairness,
  accountability and transparency. pp. 77--91. PMLR (2018)

\bibitem{dong2017class}
Dong, Q., Gong, S., Zhu, X.: Class rectification hard mining for imbalanced
  deep learning. In: Proceedings of the IEEE/CVF International Conference on
  Computer Vision. pp. 1851--1860 (2017)

\bibitem{gebru2018datasheets}
Gebru, T., Morgenstern, J., Vecchione, B., Vaughan, J.W., Wallach, H.,
  Daum{\'e}~III, H., Crawford, K.: Datasheets for datasets. arXiv preprint
  arXiv:1803.09010  (2018)

\bibitem{golle2008machine}
Golle, P.: Machine learning attacks against the asirra captcha. In: Proceedings
  of the 15th ACM conference on Computer and communications security. pp.
  535--542 (2008)

\bibitem{gong2019diversity}
Gong, Z., Zhong, P., Hu, W.: Diversity in machine learning. IEEE Access
  \textbf{7},  64323--64350 (2019)

\bibitem{he2016deep}
He, K., Zhang, X., Ren, S., Sun, J.: Deep residual learning for image
  recognition. In: Proceedings of the IEEE/CVF Conference on Computer Vision
  and Pattern Recognition. pp. 770--778 (2016)

\bibitem{jo2020lessons}
Jo, E.S., Gebru, T.: Lessons from archives: Strategies for collecting
  sociocultural data in machine learning. In: Proceedings of the 2020
  Conference on Fairness, Accountability, and Transparency. pp. 306--316 (2020)

\bibitem{kadambi2021achieving}
Kadambi, A.: Achieving fairness in medical devices. Science
  \textbf{372}(6537),  30--31 (2021)

\bibitem{karkkainen2021fairface}
Karkkainen, K., Joo, J.: Fairface: Face attribute dataset for balanced race,
  gender, and age for bias measurement and mitigation. In: Proceedings of the
  IEEE/CVF Winter Conference on Applications of Computer Vision. pp. 1548--1558
  (2021)

\bibitem{larrazabal2020gender}
Larrazabal, A.J., Nieto, N., Peterson, V., Milone, D.H., Ferrante, E.: Gender
  imbalance in medical imaging datasets produces biased classifiers for
  computer-aided diagnosis. Proceedings of the National Academy of Sciences
  \textbf{117}(23),  12592--12594 (2020)

\bibitem{li2019repair}
Li, Y., Vasconcelos, N.: Repair: Removing representation bias by dataset
  resampling. In: Proceedings of the IEEE/CVF Conference on Computer Vision and
  Pattern Recognition. pp. 9572--9581 (2019)

\bibitem{loshchilov2017decoupled}
Loshchilov, I., Hutter, F.: Decoupled weight decay regularization. arXiv
  preprint arXiv:1711.05101  (2017)

\bibitem{mehrabi2021survey}
Mehrabi, N., Morstatter, F., Saxena, N., Lerman, K., Galstyan, A.: A survey on
  bias and fairness in machine learning. ACM Computing Surveys (CSUR)
  \textbf{54}(6),  1--35 (2021)

\bibitem{paszke2019pytorch}
Paszke, A., Gross, S., Massa, F., Lerer, A., Bradbury, J., Chanan, G., Killeen,
  T., Lin, Z., Gimelshein, N., Antiga, L., et~al.: Pytorch: An imperative
  style, high-performance deep learning library. Advances in neural information
  processing systems  \textbf{32} (2019)

\bibitem{rajpurkar2017chexnet}
Rajpurkar, P., Irvin, J., Zhu, K., Yang, B., Mehta, H., Duan, T., Ding, D.,
  Bagul, A., Langlotz, C., Shpanskaya, K., et~al.: Chexnet: Radiologist-level
  pneumonia detection on chest x-rays with deep learning. arXiv preprint
  arXiv:1711.05225  (2017)

\bibitem{ryu2017inclusivefacenet}
Ryu, H.J., Adam, H., Mitchell, M.: Inclusivefacenet: Improving face attribute
  detection with race and gender diversity. arXiv preprint arXiv:1712.00193
  (2017)

\bibitem{yao2020estimation}
Yao, Y., Yu, H., Mu, J., Li, J., Pu, H.: Estimation of the gender ratio of
  chickens based on computer vision: Dataset and exploration. Entropy
  \textbf{22}(7), ~719 (2020)

\bibitem{zhang2017age}
Zhang, Z., Song, Y., Qi, H.: Age progression/regression by conditional
  adversarial autoencoder. In: Proceedings of the IEEE/CVF Conference on
  Computer Vision and Pattern Recognition. pp. 5810--5818 (2017)

\end{thebibliography}

\end{document}


\pagestyle{headings}
\mainmatter
\def\ECCVSubNumber{1531}  

\title{MIME: Minority Inclusion for Majority Group Enhancement of AI Performance} 

\titlerunning{Minority Inclusion for Majority Enhancement of AI Performance}
%

\author{Pradyumna Chari\inst{1} 
\and
Yunhao Ba\inst{1} 
\and
Shreeram Athreya\inst{1}
\and
Achuta Kadambi\inst{1,2}
}

\authorrunning{P. Chari et al.}

\institute{Department of Electrical and Computer Engineering, UCLA\\ \and
Department of Computer Science, UCLA\\
\email{\{pradyumnac,yhba,shreeram\}@ucla.edu}, \email{achuta@ee.ucla.edu}}
\maketitle

\section*{Supplementary Contents}
\noindent This supplement is organized as follows:
\begin{itemize}
    \item Section~\ref{sec:proof1} contains the proof for Theorem 1. 
    \item Section~\ref{sec:proof2} contains the proof for Theorem 2.
    \item Section~\ref{sec:proof3} contains the proof for Theorem 3.
    \item Section~\ref{sec:beyond_1D} discusses MIME existence beyond 1D.
    \item Section~\ref{sec:histograms} describes further details for the feature space analysis.
    \item Section~\ref{sec:implementation} contains implementation details across the six datasets.
    \item Section~\ref{sec:additional_results} describes additional secondary analysis.
    \item Section~\ref{sec:hardmining} describes our implementation of the hard mining comparison.
    \item Section~\ref{sec:code} describes our code.
    \item Section~\ref{sec:negative} discusses potential negative ethical impacts of this work.
\end{itemize}

\section{Proof for Theorem 1}
\label{sec:proof1}
\noindent We consider the one-dimensional linear classifier setting, trained using the Perceptron algorithm. Given any $x \in \mathbb{R}$, the classifier evaluates an output $y$ given by,
\begin{equation}
    y=wx+b,
\end{equation}
where $w,b\in \mathbb{R}$. The decision threshold in this case is at $y=0$. For simplification, we reduce the redundant parameter, as follows:
\begin{equation}
    y'=x+b'.
\end{equation}
Note that the decision threshold is unaffected by this conversion. For notational simplicity, we use $y=y'$ and $b=b'$ here onward. We consider the perceptron decision and update rule, modified for our case. That is, for any training sample $(x_i,y_i)$, the predicted output is given by,
\begin{equation}
    \hat{y}_i=\frac{sign(x_i+b)+3}{2},
\end{equation}
where $sign(\cdot)$ is the sign function. Readers will notice the unconventional form of this decision rule. The additional terms map the conventional perceptron labels in $\{-1,1\}$ to our chosen labels $\{1,2\}$ respectively.

\noindent For an appropriately chosen learning rate $\gamma$, the parameter update rule for this setting is given by:

\begin{equation}
  b\leftarrow
  \begin{cases}
    b+\gamma, & \text{if } \hat{y}_i \neq y_i \textrm{ and } y_i=2 \\
    b-\gamma, & \text{if } \hat{y}_i \neq y_i \textrm{ and } y_i=1
  \end{cases}.
\end{equation}

\noindent Let $\mathbf{h}_{\textrm{ideal}}\triangleq [1,\,b_{\textrm{ideal}}]^T$ denote the ideal decision hyperplane. Under the current assumption of no domain gap, it can be shown that this ideal hyperplane is located at $x=d_{\textrm{ideal}}$ such that,
\begin{equation}
    \begin{split}
        p^{\textrm{minor}}_1 (d_{\textrm{ideal}})&=p^{\textrm{minor}}_2 (d_{\textrm{ideal}})\\
        p^{\textrm{major}}_1 (d_{\textrm{ideal}})&=p^{\textrm{major}}_2 (d_{\textrm{ideal}}).
    \end{split}
\end{equation}
This also implies that $b_{\textrm{ideal}}=-d_{\textrm{ideal}}$. Now, consider an initial training set of $K-1$ samples from the majority group, $\mathcal{D}_{K-1}^{\textrm{major}}$. A decision hyperplane $\mathbf{h}_{K-1}$ is learnt from these samples. Then, without loss of generality, we can assume that,
\begin{equation}
    d_{K-1}=d_{\textrm{ideal}}+\Delta.
\end{equation}
That is, the real hyperplane $\mathbf{h}_{K-1}$ is non-ideally located closer to the positve class ($y=2$) than $\mathbf{h}_{\textrm{ideal}}$. $\Delta$ is a small positive value representing the error in the learnt decision hyperplane. Consider that the $K$-th sample is drawn from the majority group $x_K^{\textrm{major}}$. Recall that parameter updates for the Perceptron algorithm take place only in the event of incorrect label estimation $\hat{y}_K\neq y_K$. If we denote the change in the parameter $b$ due to this sample as $\Delta b$, then three cases exist:
\begin{enumerate}
    \item Sample from class 2 is classified as belonging to class 1 such that\\
    $x_K^{\textrm{major}}\sim p^{\textrm{minor}}_2 (x),\textrm{ }x_K^{\textrm{major}}<d_{\textrm{ideal}}-\Delta$. Associated $\Delta b=+\gamma$.

    \item Sample from class 2 is classified as belonging to class 1 such that\\ 
    $x_K^{\textrm{major}}\sim p^{\textrm{minor}}_2 (x),\textrm{ }d_{\textrm{ideal}}-\Delta \leq x_K^{\textrm{major}}<d_{\textrm{ideal}}+\Delta$. Associated $\Delta b=+\gamma$. 
    
    \item Sample from class 1 classified as belonging to class 2 such that\\ 
    $x_K^{\textrm{major}}\sim p^{\textrm{minor}}_1 (x),\textrm{ }x_K^{\textrm{major}}\geq d_{\textrm{ideal}}+\Delta$. Associated $\Delta b=-\gamma$. 
\end{enumerate}

\noindent Let the expected change in $b$ due to one majority group sample be denoted as $\Delta b^{\textrm{major}}$. $\Delta d^{\textrm{major}}$ is similarly defined for the expected change in $d$. Then, the following holds true:
\begin{equation}
    \Delta b^{\textrm{major}}=\underset{x_K^{\textrm{major}}}{\mathbb{E}}\left[\Delta b\right].
\end{equation}
Writing out the expectation over all three cases,
\begin{multline}
    \Delta b^{\textrm{major}}=\gamma\int^{d_{\textrm{ideal}}-\Delta}_{x=-\infty}p^{\textrm{major}}_2 (x)dx
    +\gamma\int^{d_{\textrm{ideal}}+\Delta}_{x=d_{\textrm{ideal}}-\Delta}p^{\textrm{major}}_2 (x)dx\\
    -\gamma\int^{+\infty}_{x=d_{\textrm{ideal}}+\Delta}p^{\textrm{major}}_1 (x)dx.  
    \label{eq:change}
\end{multline}

\noindent Similar expressions can be identified if the $K$-th sample is drawn from the minority group. Under the assumption that the mixture models under consideration are symmetric Gaussian mixture models,

\begin{equation}
    \int^{d_{\textrm{ideal}}-\Delta}_{x=-\infty}p^{\textrm{major}}_2 (x)dx =\int^{+\infty}_{x=d_{\textrm{ideal}}+\Delta}p^{\textrm{major}}_1 (x)dx.
    \label{eq:equal_tails}
\end{equation}
Then, using Equation~\ref{eq:change} and Equation~\ref{eq:equal_tails},
\begin{equation}
     \Delta b^{\textrm{major}}=\gamma\int^{d_{\textrm{ideal}}+\Delta}_{x=d_{\textrm{ideal}}-\Delta}p^{\textrm{major}}_2 (x)dx.  
     \label{eq:important_part}
\end{equation}
The region between $x=d_{\textrm{ideal}}-\Delta$ and $d_{\textrm{ideal}}+\Delta$ determines the expected change in the classification parameter. If $\Delta$ is small enough,
$\Delta b^{\textrm{major}}\approx 2\gamma p^{\textrm{major}}_2 (d_{\textrm{ideal}})\Delta$. Similarly, $\Delta b^{\textrm{minor}}\approx 2\gamma p^{\textrm{minor}}_2 (d_{\textrm{ideal}})\Delta$.

\noindent We now identify a sufficient condition where $p^{\textrm{minor}}_2 (x)>p^{\textrm{major}}_2 (x)$ for $-\Delta\leq x\leq \Delta$, given that the overlaps satisfy the condition $O_{\textrm{minor}}>O_{\textrm{major}}$, as defined in the main text.
\noindent Under the GMM assumption,
\begin{equation}
    p^{\textrm{major}}_2 (x)= \frac{1}{\sqrt{2\pi(\sigma^{\textrm{major}}_{2})^2}}exp\left(-\frac{(x-\mu^{\textrm{major}}_{2})^2}{2(\sigma^{\textrm{major}}_{2})^2}\right).
\end{equation}
A similar expression exists for the minority group distribution as well. We wish to find the intersection point for the majority and minority distributions, that is $p^{\textrm{major}}_2 (x)=p^{\textrm{major}}_1 (x)$ for some $x$. This expression reduces to,
\begin{equation}
    \frac{\left(x-\mu^{\textrm{major}}_{2}\right)^2}{{\sigma^{\textrm{major}}}^2}-\frac{\left(x-\mu^{\textrm{minor}}_{2}\right)^2}{{\sigma^{\textrm{minor}}}^2}=2ln\bigg\lvert\frac{\sigma^{\textrm{minor}}}{\sigma^{\textrm{major}}}\bigg\lvert.
\end{equation}

\noindent We want to ensure that this intersection point occurs for an $x>d_{\textrm{ideal}}$. This sets up a hyperbolic equation for the condition. For our purposes of proving existence, we qualitatively note that if the majority group variance is not very large (meaning the likelihood of sampling at the ideal hyperplane is low for the majority group), and the minority group variance is not very large (such that it does not tend close to a uniform distribution), $p^{\textrm{minor}}_2 (x)>p^{\textrm{major}}_2 (x)$. Then,
\begin{equation}
    \Delta b^{\textrm{minor}}>\Delta b^{\textrm{major}}.
\end{equation}
\begin{equation}
    \Delta d^{\textrm{minor}}<\Delta d^{\textrm{major}}<0.
    \label{eq:d_relation}
\end{equation}

\noindent Our final task is to relate the expected change in the decision hyperplane over a choice of training sets $\mathcal{D}_K^{+}$ and $\mathcal{D}_K^{-}$, with associated learnt hyperplanes $\mathbf{h}^+_K$ and $\mathbf{h}^-_K$. As a reminder, \begin{equation}
\begin{split}
    \mathcal{D}_K^{+} &= \{ \mathcal{D}_{K-1}^{\textrm{major}}, x_K^{\textrm{major}} \} \\ 
      \mathcal{D}_K^{-} &= \{ \mathcal{D}_{K-1}^{\textrm{major}}, x_K^{\textrm{minor}} \}, 
\end{split}
\end{equation}

\noindent Consider a general training setting, where we use minibatches of size $M>1$, over multiple epochs. Then, any minibatch containing the $K$-th sample can be split into the $K$-th sample and a random subset of $M-1$ samples from $\mathcal{D}_{K-1}^{\textrm{major}}$. Therefore, on average, the only difference to the sample updates would be due to the contributions of the $K$-th sample. This brings us to our final observations,
\begin{equation}
\begin{split}
    \underset{x_K^{\textrm{minor}}}{\mathbb{E}}[d^{+}_K] &= d^{\textrm{major}}_{K-1}+\Delta d^{\textrm{major}}\\
    \underset{x_K^{\textrm{minor}}}{\mathbb{E}}[d^{-}_K] &= d^{\textrm{major}}_{K-1}+\Delta d^{\textrm{minor}}.
\end{split}
\label{eq:change2}
\end{equation}
From Equations~\ref{eq:d_relation} and~\ref{eq:change2},
\begin{equation}
    \underset{x_K^{\textrm{minor}}}{\mathbb{E}}[d^{-}_K]<\underset{x_K^{\textrm{minor}}}{\mathbb{E}}[d^{+}_K],\textrm{ and}
\end{equation}
\begin{equation}
    \underset{x_K^{\textrm{minor}}}{\mathbb{E}}[\lvert d_{\textrm{ideal}}-d^{-}_{K}\lvert]<\underset{x_K^{\textrm{minor}}}{\mathbb{E}}[\lvert d_{\textrm{ideal}}-d^{+}_{K}\lvert].
\end{equation}
The above holds for small enough $\gamma$. Since we know the relationship between the decision hyperplane $\mathbf{h}$ and the associated $d$ in our setting, the following equations hold true:
\begin{equation}
    \underset{x_K^{\textrm{minor}}}{\mathbb{E}}\lVert\mathbf{h}_{\textrm{ideal}}-\mathbf{h}^-_K\lVert<\underset{x_K^{\textrm{minor}}}{\mathbb{E}}\lVert\mathbf{h}_{\textrm{ideal}}-\mathbf{h}^+_K\lVert,
\end{equation}
\begin{equation}
    \underset{x_K^{\textrm{minor}}}{\mathbb{E}} \mathcal{P}^{\textrm{major}} \big\{ \mathbf{h}^-_K \big\} <\underset{x_K^{\textrm{major}}}{\mathbb{E}} \mathcal{P}^{\textrm{major}} \big\{ \mathbf{h}^+_K \big\}. \textrm{ }\blacksquare
\end{equation}

\section{Proof for Theorem 2}
\label{sec:proof2}
\noindent We follow a similar approach as in Theorem 1.
Let $\mathbf{h}^{\textrm{major}}_{\textrm{ideal}}\triangleq [1,\,b^{\textrm{major}}_{\textrm{ideal}}]^T$ denote the ideal decision hyperplane for the majority group. Let $\mathbf{h}^{\textrm{minor}}_{\textrm{ideal}}\triangleq [1,\,b^{\textrm{minor}}_{\textrm{ideal}}]^T$ denote the ideal decision hyperplane for the minority group. Then, the ideal hyperplanes are located at $x=d^{\textrm{major}}_{\textrm{ideal}}$ and $x=d^{\textrm{minor}}_{\textrm{ideal}}$ respectively such that,
\begin{equation}
    \begin{split}
        p^{\textrm{minor}}_1 (d^{\textrm{minor}}_{\textrm{ideal}})&=p^{\textrm{minor}}_2 (d^{\textrm{minor}}_{\textrm{ideal}})\\
        p^{\textrm{major}}_1 (d^{\textrm{major}}_{\textrm{ideal}})&=p^{\textrm{major}}_2 (d^{\textrm{major}}_{\textrm{ideal}}).
    \end{split}
    \label{eq:ideal_hyperplanes}
\end{equation}
This implies that $b^{\textrm{major}}_{\textrm{ideal}}=-d^{\textrm{major}}_{\textrm{ideal}}$ and $b^{\textrm{minor}}_{\textrm{ideal}}=-d^{\textrm{minor}}_{\textrm{ideal}}$. Consider an initial training set of $K-1$ samples from the majority group, $\mathcal{D}_{K-1}^{\textrm{major}}$. Then, without loss of generality, we can assume that $d_{K-1}=d^{\textrm{major}}_{\textrm{ideal}}+\Delta$,
where $\Delta>0$. Additionally, we consider the existence of domain gap in this case, that is, $d^{\textrm{minor}}_{\textrm{ideal}}=d^{\textrm{major}}_{\textrm{ideal}}+\delta$.

\noindent Let $\delta<\Delta$. Similar to the setting in Theorem 1 (Equation~\ref{eq:change}), we can set up the equation for expected parameter change in the case of the majority and minority groups as follows:
\begin{multline}
     \Delta b^{\textrm{major}}=\gamma\int^{d^{\textrm{major}}_{\textrm{ideal}}-\Delta}_{x=-\infty}p^{\textrm{major}}_2 (x)dx
    +\gamma\int^{d^{\textrm{major}}_{\textrm{ideal}}+\Delta}_{x=d^{\textrm{major}}_{\textrm{ideal}}-\Delta}p^{\textrm{major}}_2 (x)dx\\
    -\gamma\int^{+\infty}_{x=d^{\textrm{major}}_{\textrm{ideal}}+\Delta}p^{\textrm{major}}_1 (x)dx.  
    \label{eq:change_major2}
\end{multline}

\begin{multline}
     \Delta b^{\textrm{minor}}=\gamma\int^{d^{\textrm{minor}}_{\textrm{ideal}}-(\Delta-\delta)}_{x=-\infty}p^{\textrm{minor}}_2 (x)dx
    +\gamma\int^{d^{\textrm{minor}}_{\textrm{ideal}}+(\Delta-\delta)}_{x=d^{\textrm{minor}}_{\textrm{ideal}}-(\Delta-\delta)}p^{\textrm{minor}}_2 (x)dx\\
    -\gamma\int^{+\infty}_{x=d^{\textrm{minor}}_{\textrm{ideal}}+(\Delta-\delta)}p^{\textrm{minor}}_1 (x)dx.  
    \label{eq:change_minor2}
\end{multline}

\noindent Under the assumption that the mixture models under consideration are symmetric Gaussian mixture models,

\begin{equation}
     \Delta b^{\textrm{major}}=\gamma\int^{d^{\textrm{major}}_{\textrm{ideal}}+\Delta}_{x=d^{\textrm{major}}_{\textrm{ideal}}-\Delta}p^{\textrm{major}}_2 (x)dx,  
\end{equation}

\begin{equation}
     \Delta b^{\textrm{minor}}=\gamma\int^{d^{\textrm{minor}}_{\textrm{ideal}}+(\Delta-\delta)}_{x=d^{\textrm{minor}}_{\textrm{ideal}}-(\Delta-\delta)}p^{\textrm{minor}}_2 (x)dx.  
\end{equation}

\noindent If $\Delta+\lvert \delta \lvert$ is small enough,
\begin{equation}
    \Delta b^{\textrm{major}}\approx 2\gamma p^{\textrm{major}}_2 (d^{\textrm{major}}_{\textrm{ideal}})\Delta,
\end{equation}
\begin{equation}
    \Delta b^{\textrm{minor}}\approx 2\gamma p^{\textrm{minor}}_2 (d^{\textrm{minor}}_{\textrm{ideal}})(\Delta-\delta).
\end{equation}

\noindent By establishing the same conditions on group class variances as Theorem 1, we know that $ p^{\textrm{minor}}_2 (d^{\textrm{minor}}_{\textrm{ideal}})> p^{\textrm{major}}_2 (d^{\textrm{major}}_{\textrm{ideal}})$. We now identify conditions under which $\Delta b^{\textrm{minor}}>\Delta b^{\textrm{major}}$.

\paragraph{Case 1 - $\boldsymbol{\delta<}\mathbf{0}$:} 

Under the same conditions as Theorem 1, 
$(\Delta-\delta)>\Delta$, and $p^{\textrm{minor}}_2 (d^{\textrm{minor}}_{\textrm{ideal}})>p^{\textrm{major}}_2 (d^{\textrm{major}}_{\textrm{ideal}})$. Therefore,
\begin{equation}
    \Delta b^{\textrm{minor}}>\Delta b^{\textrm{major}}.
\end{equation}

\paragraph{Case 2 - $\boldsymbol{\delta>}\mathbf{0}$:}
\begin{equation}
    \Delta b^{\textrm{major}}\approx 2\gamma p^{\textrm{major}}_2 (d^{\textrm{major}}_{\textrm{ideal}})\Delta,
\end{equation}
\begin{equation}
    \Delta b^{\textrm{minor}}\approx 2\gamma p^{\textrm{minor}}_2 (d^{\textrm{minor}}_{\textrm{ideal}})(\Delta-\delta).
\end{equation}
For $\Delta b^{\textrm{minor}}>\Delta b^{\textrm{major}}$,
\begin{equation}
     p^{\textrm{minor}}_2 (d^{\textrm{minor}}_{\textrm{ideal}})(\Delta-\delta)>p^{\textrm{major}}_2 (d^{\textrm{major}}_{\textrm{ideal}})\Delta.
     \label{eq:import}
\end{equation}
Rearranging Equation~\ref{eq:import},
\begin{equation}
    \frac{p^{\textrm{major}}_2 (d^{\textrm{major}}_{\textrm{ideal}})}{p^{\textrm{minor}}_2 (d^{\textrm{minor}}_{\textrm{ideal}})}<\left(1-\frac{\delta}{\Delta}\right).
    \label{eq:ratio}
\end{equation}
Given the definitions of the majority and minority groups,
\begin{equation}
    p^{\textrm{major}}_2 (d^{\textrm{major}}_{\textrm{ideal}})<p^{\textrm{minor}}_2 (d^{\textrm{minor}}_{\textrm{ideal}}),
\end{equation}
\begin{equation}
    O_{\textrm{major}}<O_{\textrm{minor}}.
\end{equation}
Since all four of these terms depend only on the means and variances of the Gaussian components, we can write,
\begin{equation}
    \frac{O_{\textrm{major}}}{O_{\textrm{minor}}}=\frac{p^{\textrm{major}}_2 (d^{\textrm{major}}_{\textrm{ideal}})}{p^{\textrm{minor}}_2 (d^{\textrm{minor}}_{\textrm{ideal}})}f,
    \label{eq:ratio_relation}
\end{equation}
where $f$ is a positive scalar constant that depends only on the component means and variances. From Equations~\ref{eq:ratio} and~\ref{eq:ratio_relation},
\begin{equation}
    \frac{O_{\textrm{major}}}{O_{\textrm{minor}}}<\left(1-\frac{\delta}{\Delta}\right)f.
\end{equation}
\noindent This proves the conditions in the theorem. Theorem 1 can now be used to show the existence of the MIME effect in the presence of domain gap, for these conditions. $\blacksquare$
  
\paragraph{A Note on the Theorems:}
\noindent Theorems 1 and 2 are existence theorems. That is, they show that there exist certain conditions under which the MIME effect can be observed. The theorems make these arguments based on the `usefulness' of points close to the ideal hyperplane. The direct metric of correlation is the likelihood for a particular distribution to sample at the ideal hyperplane. However, since this cannot be easily measured in practice, we set up our proofs in terms of a correlated metric: the overlap.

\section{Proof for Theorem 3}
\label{sec:proof3}
This Theorem considers distributions with general prior distributions. Therefore, for the majority group, let
\begin{equation}
\begin{split}
    {p}^{\textrm{major}'}_2(x)&=\pi^{\textrm{major}}p^{\textrm{major}}_2(x),\\
    {p}^{\textrm{major}'}_q(x)&=(1-\pi^{\textrm{major}})p^{\textrm{major}}_1(x).
\end{split}
\end{equation}
Similar definitions are made for the minority group as well. Then, assuming $d_{K-1}=d^{\textrm{major}}_{\textrm{ideal}}+\Delta$, $\Delta>0$ (similar to Theorem 2), and $\delta=0$ (for now), and drawing from Equation~\ref{eq:change}), we can set up the equation for expected parameter change in the case of the majority group as follows:
\begin{equation}
\begin{split}
    \Delta b^{\textrm{major}}&=\gamma\int^{d_{\textrm{ideal}}+\Delta}_{x=-\infty}p^{\textrm{major}'}_2 (x)dx
    -\gamma\int^{+\infty}_{x=d_{\textrm{ideal}}+\Delta}p^{\textrm{major}'}_1 (x)dx\\ &=T^{\textrm{major}}(d_{\textrm{ideal}}+\Delta).  
\end{split}
    \label{eq:change3}
\end{equation}
A similar expression holds true for the minority group. Then, if $T^{\textrm{major}}(d_{\textrm{ideal}}+\Delta)<T^{\textrm{minor}}(d_{\textrm{ideal}}+\Delta)$, the MIME effect will hold true.

Similarly, if $d_{K-1}=d^{\textrm{major}}_{\textrm{ideal}}-\Delta$, $\Delta>0$,
\begin{equation}
\begin{split}
    \Delta b^{\textrm{major}}&=-\gamma\int^{d_{\textrm{ideal}}-\Delta}_{x=-\infty}p^{\textrm{major}'}_2 (x)dx
    +\gamma\int^{+\infty}_{x=d_{\textrm{ideal}}-\Delta}p^{\textrm{major}'}_1 (x)dx\\ &=-T^{\textrm{major}}(d_{\textrm{ideal}}-\Delta).  
\end{split}
    \label{eq:change4}
\end{equation}
Then, if $-T^{\textrm{major}}(d_{\textrm{ideal}}-\Delta)<-T^{\textrm{minor}}(d_{\textrm{ideal}}-\Delta)$, the MIME effect will hold true.

Combining the two expressions, for a sufficient existence condition, we get,

 \begin{multline}
     \textrm{min}\left\{ T^{\textrm{minor}}(d_{\textrm{ideal}}+\Delta), -T^{\textrm{minor}}(d_{\textrm{ideal}}-\Delta)\right\}>\\
     \textrm{max}\left\{ T^{\textrm{major}}(d_{\textrm{ideal}}+\Delta), -T^{\textrm{major}}(d_{\textrm{ideal}}-\Delta)\right\}.
     \label{eq:th3supp}
 \end{multline}

 This completes the proof. $\blacksquare$
 
 \noindent Note that the existence proof for Theorem 3 ignores the effect of domain gap $\delta$, in the interest of readability and brevity. A very similar existence proof can be established with domain gap. We omit the derivation and provide the final condition below (under the constraints on $\delta$ and $\Delta$ as in Theorem 2, and using the same notation):
 
\begin{multline}
     \textrm{min}\left\{ T^{\textrm{minor}}(d^{\textrm{major}}_{\textrm{ideal}}+\Delta), -T^{\textrm{minor}}(d^{\textrm{major}}_{\textrm{ideal}}-\Delta)\right\}>\\
     \textrm{max}\left\{ T^{\textrm{major}}(d^{\textrm{major}}_{\textrm{ideal}}+\Delta), -T^{\textrm{major}}(d^{\textrm{major}}_{\textrm{ideal}}-\Delta)\right\}.
     \label{eq:th3suppdelta}
 \end{multline}

\section{MIME Existence Beyond 1D Settings}
\label{sec:beyond_1D}
\noindent Consider $\mathbf{x}\in \mathbb{R}^n$. The perceptron decisions are based on the metric $y=\mathbf{w}^T\mathbf{x}+b$, where $\mathbf{w}\in \mathbb{R}^n$, and $y,b\in \mathbb{R}$. Similar to Theorem 1, we consider the perceptron decision and update rule. That is, for any training sample $(\mathbf{x}_i,y_i)$, the predicted label is given by,
\begin{equation}
    \hat{y}_i=\frac{sign(\mathbf{w}^T\mathbf{x}_i+b)+3}{2}.
\end{equation}
We can rewrite this in terms of a single decision hyperplane by defining $\mathbf{\Tilde{w}}=[\mathbf{w}^T\,\,b]^T$ and $\mathbf{\Tilde{x}}=[\mathbf{x}^T\,\,1]^T$. For a small learning rate $\gamma$, the updated decision rule becomes,
\begin{equation}
    \hat{\Tilde{y}}_i=\frac{sign(\mathbf{\Tilde{w}}^T{\mathbf{\Tilde{x}}_i})+3}{2}.
    \label{eq:bias_consume}
\end{equation}

\begin{equation}
  \mathbf{\Tilde{w}}\leftarrow
  \begin{cases}
    \mathbf{\Tilde{w}} +\gamma \mathbf{\Tilde{x}}_i, & \text{if } \hat{y}_i \neq y_i \textrm{ and } y_i=2 \\
    \mathbf{\Tilde{w}} -\gamma \mathbf{\Tilde{x}}_i, & \text{if } \hat{y}_i \neq y_i \textrm{ and } y_i=1
  \end{cases}.
\end{equation}

\noindent We now refer to the hyperplane $\mathbf{\Tilde{w}}$ as the decision hyperplane.
Let $\mathbf{h}_{\textrm{ideal}}$ be the ideal decision hyperplane. 
In this setting, any domain gap $\delta$ or error in real hyperplane estimation $\Delta$ manifests as a direction/angle error in the hyperplane normal vector (since the bias term $b$ is subsumed in the hyperplane). The updates change the normal vector of the hyperplane through a linear combination with the sample $\mathbf{\Tilde{x}}_i$, scaled by the learning rate $\gamma$. 

\begin{figure}[ht]
    \centering
    \includegraphics[width=0.6\textwidth]{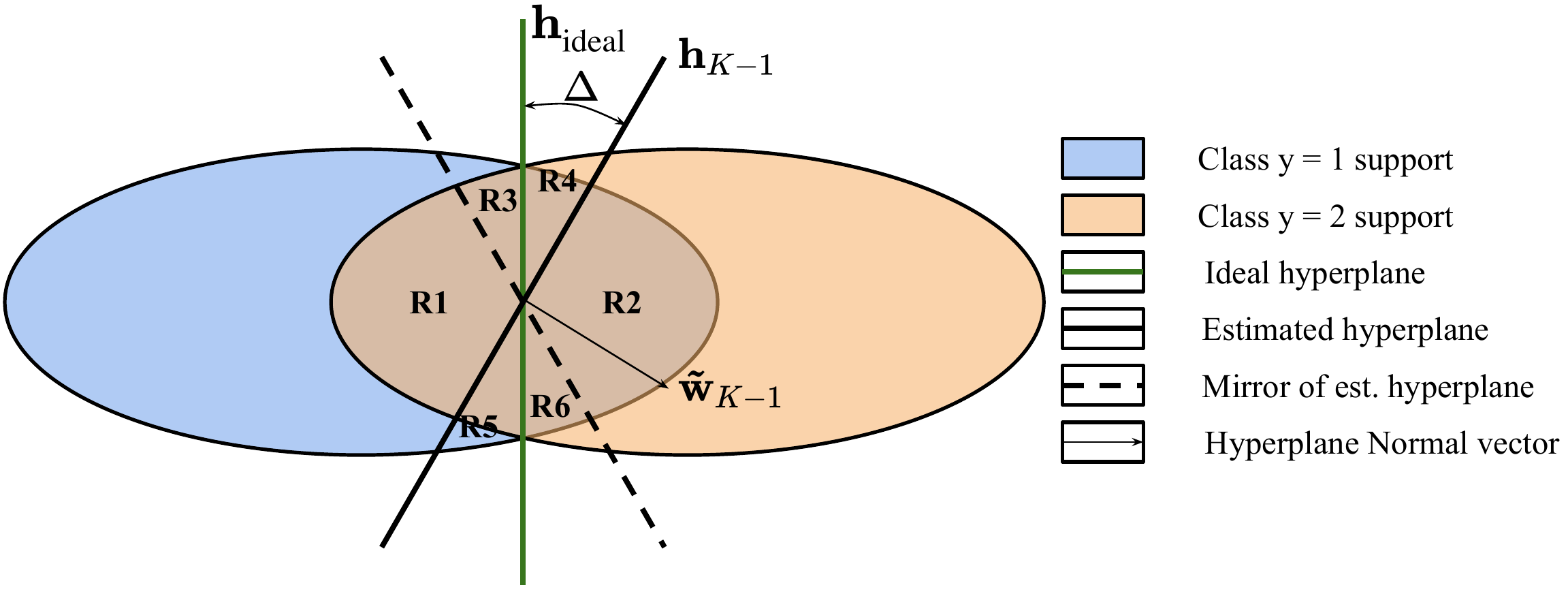}
    \caption{\textbf{The MIME effect holds in a multidimensional setting as well.} We show the support for the two finite distributions. Weight vector updates arising out of samples from regions R3, R4, R5 and R6 lead to an update with a large vertical (corrective) component (favorable update). Updates arising out of regions R1 and R2 result in an overall update in the horizontal direction (unfavorable update).}
    \label{fig:2D}
\end{figure}

\noindent We now provide a qualitative description for the existence of the MIME effect, in terms of the likelihood of a favorable update to $\mathbf{\Tilde{w}}$. We consider a simplified 2D case with symmetric distributions and $\delta=0$. A finite support is assumed for the majority and minority groups, for ease of understanding. Consider that the bias term $b$ is known, and only the hyperplane direction is to be refined. Again, we denote the hyperplane from our finite training set $\mathcal{D}_{K-1}^{\textrm{major}}$ as $\mathbf{h}_{K-1}$. The error $\Delta$ in this case is now the angular error between the normals for $\mathbf{h}_{\textrm{ideal}}$ and $\mathbf{h}_{K-1}$. Figure~\ref{fig:2D} indicates this setting. The learnt hyperplane $\mathbf{\Tilde{w}}_{K-1}$ is shown as a black solid line. The black dashed line represents the mirror image of the learnt hyperplane, defined for aid in simplification. Recall that updates to the weight vector take place on misclassification. On average, the updates due to samples in regions R1 for ($y=2$) and R2 (for $y=1$) lead to a net horizontal (leftward) weight update. This is an unfavorable update that increases $\Delta$. Therefore, the favorable updates on average are from regions R3 and R4 for $y=2$, and R5 and R6 for $y=1$. This is a net update with large vertical (upward) update. This is a favorable update that decreases $\Delta$. These regions are described based on the small angular deviation $\Delta$. Since the distributions have finite support along the direction parallel to the ideal hyperplane (vertical direction in Figure~\ref{fig:2D}), the requirement again reduces to greater likelihood of sampling close to the ideal hyperplane (similar to Theorems 1 and 2), since $\Delta$ is small. That is, distributions that sample close to the ideal hyperplane with greater probability have a greater expected likelihood of a favorable update. Under similar conditions as Theorem 1, MIME effect holds in this case. 

\noindent The extension to include the bias term $b$ is straightforward. We follow the setting in Equation~\ref{eq:bias_consume} and subsume the bias as part of the weights. In this case, $\Delta$ includes the error in both the hyperplane normal direction as well as the bias. Extensions to greater number of dimensions can be done using the same arguments. Additionally, domain gap can also be introduced. We omit explicit mathematical expressions in the interest of brevity, and since our goal here is to establish existence.

\noindent

\section{Feature Space Analysis}
\label{sec:histograms}

\paragraph{Constructing the Projected Feature Histograms:}
Let $f$ denote a feature vector, in the penultimate layer of a classification neural network. For example, in the case of ResNet-34~\cite{he2016deep}, $\mathbf{f}\in \mathbb{R}^{512}$. Similarly, let $\mathbf{w}$ be the final layer weights. In the case of multiple final layer hyperplanes, we choose any one of the hyperplanes (since for the two class classification task, the two projected variables are correlated when trained against the cross entropy loss for 2 classes). Then, we define $x\in \mathbb{R}$ as,
\begin{equation}
    x = \mathbf{w}^T\mathbf{f}.
\end{equation}
Classification decisions are made solely on the basis of the projected variable $x$. Therefore, we analyze the histogram distributions for $x$. Practically, for each dataset, we use the best performing (in terms of majority group performance) model trained using a minority training fraction ($\beta$) of 0.5. This is chosen in order to obtain histograms of $x$ for all four distributions -- the two task classes for both the majority and minority groups. The histograms are created using the test set samples.

\paragraph{Estimating the Overlap:}
The overlap is estimated from the histograms, using the following Python code snippet:

\begin{python}
    def histogram_intersection(h1, h2, bins):
        #INPUTS:
        #h1, h2: normalized histograms
        #bins: number of bins in the histograms (should be equal for the two histograms)
        #OUTPUTS:
        #sm: overlap fraction
        sm = 0
        for i in range(bins):
            sm += min(h1[i], h2[i])
        return sm
\end{python}

\paragraph{Estimating the Domain Gap:}
We follow a two step process to estimate the domain gap $\delta$. First, the ideal decision hyperplanes for the majority and minority groups are estimated, using Equation~\ref{eq:ideal_hyperplanes}. We fit a fifth order polynomial to the two histograms. The central intersection point of the histograms (i.e the intersection point that lies between the means of the two classes) is then the location of the ideal decision threshold. The following Python code snippet describes this:

\begin{python}
    import numpy as np
    
    def ideal_hyperplane(h1, h2, z, ref=5):
        #INPUTS:
        #h1, h2: the two histograms, of equal length and identical bins
        #z: a list of the histogram bin centers
        #ref: Search space for the intersection of the two histograms- default is from -5 to 5
        #OUTPUT:
        #z_dec: Ideal decision threshold between the two histogram distributions
        z_dash = np.polyfit(z, h1, 5)
        f1 = np.poly1d(z_dash)
        # calculate polynomial
        z_dash = np.polyfit(z, f2, 5)
        f2 = np.poly1d(z_dash)
        new_z = np.linspace(-ref,ref,5000)
        new_f1 = f1(new_z)
        new_f2 = f2(new_z)
        id_dec = np.argmin(np.abs(new_f1-new_f2))
        z_dec = new_z[id_dec]
        return z_dec
\end{python}

\noindent The domain gap is the absolute difference between two ideal decision thresholds, for each of the two group classes. \textcolor{blue}{Figure~3} of the main paper may be referred to for a graphical visualization.

\paragraph{Notes on the Estimated Measures:}
The latent feature space analysis is not perfect. This is because the feature extraction part of the network is jointly learnt along with decision hyperplane. Histograms are plotted on the 50\% minority training ratio so as to enable a fair domain gap and overlap comparison between the two group classes. Specifically, note that we define task complexity in the main paper in terms of the minority only and majority only train sets which deviates from the setting here. The estimates for overlap and domain gap are therefore approximate correlated estimates and not exact measures. 

\paragraph{Analysis of Feature Space Gaussian-like Behavior:} 
We set up the Chi-Squared goodness of fit test on all 20 distributions under consideration (i.e. across 5 datasets and  4 distributions each per dataset). These statistics correspond to the distributions in \textcolor{blue}{Table~1} and \textcolor{blue}{Figure~4} of the main paper. Python code for testing the hypotheses is given below. The number of bins are chosen so as to ensure $\geq 5$ samples per bin on average.
\begin{python}
    from scipy.stats import chisquare
    from scipy.stats import norm
    from scipy import stats
    import pandas as pd
    
    def chi_square_stats(vals,no_bins)
        #INPUTS:
        #vals: a list of samples whose Gaussianity is to be tested
        #no_bins: number of bins (thumb rule: no_bins<len(vals)/5)
        
        tot_vals = len(vals)
        # mean and standard deviation of given data
        mean = np.mean(vals)
        std = np.std(vals)
        
        interval = []
        for i in range(1,no_bins+1):
          val = stats.norm.ppf(i/no_bins, mean, std)
          interval.append(val)
        interval.insert(0, -np.inf)
        
        lower = interval[:-1]
        upper = interval[1:]
        
        df = pd.DataFrame({'lower_limit':lower, 'upper_limit':upper})

        sorted_vals = list(sorted(vals))
        df['obs_freq'] = df.apply(lambda x:sum([i>x['lower_limit'] and i<=x['upper_limit'] for i in sorted_vals]), axis=1)
        df['exp_freq'] = tot_vals/no_bins
        
        statistic = stats.chisquare(df['obs_freq'], df['exp_freq'])
        
        p = 2    # number of parameters for 1D Gaussian
        DOF = len(df['obs_freq']) - p -1
        thresh = stats.chi2.ppf(0.95, DOF)
        
        return statistic, thresh
\end{python}

\noindent Table~\ref{tab:chi-square} highlights the evaluated chi-square statistics, as well as related parameters. Note that a lower value of the statistic is better, and the null hypothesis is not rejected when the value of the statistic is lower than the critical value. We establish the null hypothesis at a 5\% level of significance for each distribution to be that the samples are drawn from a Gaussian distribution. Distributions that are unable to reject the null hypothesis are indicated in bold. It can be seen that a large majority of the distributions indicate that the projected latent features follow a Gaussian-like distribution.

\begin{table}[ht]
    \caption{\textbf{Chi-Squared goodness of fit measures for all distributions.} Distributions with bolded values show the estimated statistics that are lower than the critical value, indicating that the null hypothesis (Gaussian distribution) cannot be rejected.}
    \label{tab:chi-square}
    \centering
    \scriptsize
    \begin{tabular}{l c c c c c c c}
    \toprule
    Dataset &\multirow{2}{*}{\begin{tabular}{@{}c@{}}No. of samples \\ per group per class\end{tabular}} & No. of Bins & Critical Value &  \multicolumn{2}{c}{Majority Group} & \multicolumn{2}{c}{Minority Group} \\
    & & & &$y=1$ & $y=2$ & $y=1$ & $y=2$  \\
    \midrule
    DS-1~\cite{karkkainen2021fairface} & 379 & 15 & 21.03 & \textbf{13.65} & 28.69 & \textbf{10.25} & \textbf{15.39}\\
    DS-2~\cite{golle2008machine} & 126 & 15 & 21.03 & \textbf{7.81} & \textbf{12.10} & \textbf{11.62} & \textbf{9.24}\\
    DS-4~\cite{rajpurkar2017chexnet} & 126 & 15 & 21.03 & \textbf{10.43} & \textbf{17.57} & \textbf{17.10} & \textbf{4.48}\\
    DS-5~\cite{blake1998uci} & 159 & 15 &21.03& \textbf{11.09} & 24.05 & \textbf{5.74} & \textbf{14.40}\\
    DS-6~\cite{yao2020estimation,zhang2017age} & 43 & 5 & 5.99 & 25.48 & \textbf{5.02} & \textbf{5.72} & \textbf{4.79}\\
    \bottomrule
    \end{tabular}
\end{table}

\section{Implementation Details}
\label{sec:implementation}
\paragraph{Analysis measures:}

For each task, we estimate the test accuracy $a_p^i(\beta)$ as a function of minority group fraction in the train set $\beta \in [0,1]$, for a trial $i\in \{1,...,N\}$, for a group class $g$ (e.g. dark skin tones). $N$ is the total number of trials. Practically, we evaluate performance for a finite set of $\beta$ values, represented by the set $B=\{0,0.1,0.2,\dots,1.0\}$. We now define the following measures.

\noindent {\textit{Average accuracy}}: For a given minority training ratio $\beta_0$, and for a given group class $g$, we define the average accuracy,
\begin{equation}
    \bar{a}_g(\beta_0)=\frac{1}{N}\sum_{i=1}^{N}a_g^i(\beta_0).
\end{equation}

\noindent {\textit{Error bounds}}: We also evaluate the \textit{trend variation} among $a_g^i(\beta)$ for various $i$. That is, we want to evaluate if across all the trials (for a particular task-dataset combination), the relative trend (of majority group performance gain) holds true. One candidate measure for this is $std_i(a_g^i(\beta))$ for each $\beta$, where $std_i(\cdot)$ is the standard deviation operator, over $i$. However, this measure will include average changes in accuracy for all splits, for a particular trial (arising out of unrelated effects such as different train or test set samples). This is unnecessary in our case. Therefore, we define our error measure $\hat{\zeta}(\beta)$ as the $\beta$-mean subtracted standard deviation. That is,

\begin{equation}
\begin{split}
        \hat{\zeta}(\beta)&=std_{i}(a_g^i(\beta)-\bar{a}_g^i),\\
        \bar{a}_g^i&=\frac{1}{|B|}\sum_{\beta \in B}a_g^i(\beta),
\end{split}
\end{equation}
where $|\cdot|$ is the cardinality operator representing the size of a set. In our graphs, we plot the average accuracy $\bar{a}_g(\beta)$ as well as the error bounds, from $\bar{a}_g(\beta)-\hat{\zeta}(\beta)$ to $\bar{a}_g(\beta)+\hat{\zeta}(\beta)$, $\forall \beta \in B$.

\paragraph{Network Architectures Used:} For all the vision-related experiments, we use the ResNet-34 architecture~\cite{he2016deep}. We only modify the output layer of the network so as to match the number of task classes (9 for Dataset 3, and 2 for all other tasks). For the Adult (Census) Dataset~\cite{blake1998uci}, we use a fully connected network with sigmoid outputs. The PyTorch~\cite{paszke2019pytorch} implementation for the model is included below.

\begin{python}
    #Model

    def act(x):
        return F.relu(x)
    
    class Network(nn.Module):
        def __init__(self,):
            super().__init__()
            self.fc1 = nn.Linear(101, 50)
            self.fc2 = nn.Linear(50, 50)
            self.fc3 = nn.Linear(50, 50)
            self.fcLast = nn.Linear(50,2)
    
        def forward(self,x):
    
            x = act(self.fc1(x))
            # x = self.b1(x)
            x = act(self.fc2(x))
            x = act(self.fc3(x))
            x = torch.sigmoid(self.fcLast(x))
            return x
\end{python}

\paragraph{General Experiment Details:}
\begin{table}[ht]
    \caption{\textbf{Training configuration and parameters for all datasets and experiments.} Parameters for each dataset are chosen so as to maximize performance.}
    \label{tab:data_details}
    \scriptsize
    \centering
    \begin{tabular}{lllllll}
    \toprule
    Dataset & DS-1~\cite{karkkainen2021fairface} & DS-2~\cite{golle2008machine} & DS-3~\cite{zhang2017age} & DS-4~\cite{rajpurkar2017chexnet} & DS-5~\cite{blake1998uci} & DS-6~\cite{yao2020estimation,zhang2017age}\\
    (Task) & (Gender) & (Species) & (Age) & (Diagnosis) & (Income) & (Gender) \\
    \midrule
    Group class & Race & Skin tone & Gender & Gender & Gender & Species\\
    Train set size & 10900 & 1500 & 7700 & 1500 & 2600 & 750\\
    Test set size (per group) & 760 & 250 & 970 & 250 & 300 & 90\\
    No. of trials & 5 & 5 & 5 & 7 & 5 & 5\\
    No. of epochs & 35 & 60 & 65 & 40 & 250 & 20\\
    Learning rate & 0.0005 & 0.0006 & 0.0006 & 0.0006 & 0.0005 & 0.0005\\
    Weight Decay & 0.08 & 0.05 & 0.05 & 0.05 & 0.08 & 0.08\\
    Input Shape/Config. & 3x100x100 & 3x256x256 & 3x100x100 & 3x256x256 & 101x1 & 3x100x100\\
    \bottomrule
    \end{tabular}
\end{table}

All experiments were carried out using PyTorch~\cite{paszke2019pytorch}. Table~\ref{tab:data_details} highlights the training parameters used for each dataset. We use different parameters for each of the datasets. These are experimentally chosen to maximize accuracy. All the models are trained using the AdamW optimizer~\cite{loshchilov2017decoupled} and the cross entropy loss. The train and test set sizes vary slightly across trials, due to different data splits. However, the train set size remains the same for all minority training ratios of a particular trial. A validation set is held out but given the small sample size of several datasets, we measure trends based on best test performance. This is to minimize the effect of sample specific performance gap in small datasets. Averaging of trends over multiple trials, and hence multiple train-test splits ensures that the trends do not overfit to a particular configuration. Each trial is run using a unique random seed. Table~\ref{tab:seeds} highlights the random seeds used for our experiments, which were randomly chosen. Input images are resized to the chosen input size for each dataset. For the Adult dataset~\cite{blake1998uci}, we use a one-hot encoding scheme for the input. The group class information is dropped from the input before passing to the network. For all the datasets, across all minority training ratios for a particular trial, we use a fixed model initialization to ensure that the changes in accuracy are completely attributable to the train data configuration.

\begin{table}[ht]
    \caption{\textbf{Random seeds used for the trials.} Seeds were chosen at random for trials to generate average trends and error bounds.}
    \label{tab:seeds}
    \centering
    \scriptsize
    \setlength{\tabcolsep}{0.01\textwidth}
    \begin{tabular}{p{0.12\textwidth}<{\centering}p{0.12\textwidth}<{\centering}p{0.12\textwidth}<{\centering}p{0.12\textwidth}<{\centering}p{0.12\textwidth}<{\centering}p{0.12\textwidth}<{\centering}p{0.12\textwidth}<{\centering\arraybackslash}} 
    \toprule
    Dataset & DS-1~\cite{karkkainen2021fairface} & DS-2~\cite{golle2008machine} & DS-3~\cite{zhang2017age} & DS-4~\cite{rajpurkar2017chexnet} & DS-5~\cite{blake1998uci} & DS-6~\cite{yao2020estimation,zhang2017age}\\
    (Task) & (Gender) & (Species) & (Age) & (Diagnosis) & (Income) & (Gender) \\
    \midrule
    \begin{tabular}{@{}c@{}}Random\\ Seeds\end{tabular} & \begin{tabular}{@{}c@{}}0, 1, 3,\\ 5, 7\end{tabular} & \begin{tabular}{@{}c@{}}21, 42, 35,\\ 28, 31\end{tabular} & \begin{tabular}{@{}c@{}}0, 55, 2,\\ 15, 6\end{tabular} & \begin{tabular}{@{}c@{}}33, 42, 24, 36\\54, 21, 28\end{tabular} & \begin{tabular}{@{}c@{}}13, 15, 17,\\ 19, 21\end{tabular} & \begin{tabular}{@{}c@{}}0, 1, 3,\\ 5, 9\end{tabular}\\
    \bottomrule
    \end{tabular}
\end{table}

\paragraph{Dataset Specific Information:}
To perform experiments on the \textbf{Pet Images Dataset}, we manually annotate light and dark fur cats and dogs from the larger dataset used in~\cite{golle2008machine}. For the age classification task on the \textbf{UTKFace Dataset}~\cite{zhang2017age}, we pre-process the age labels to match the annotation format for the FairFace dataset~\cite{karkkainen2021fairface}. For the large domain gap gender classification task using the \textbf{UTKFace and Chicken Images Datasets}~\cite{zhang2017age,yao2020estimation}, we perform gender classification over human and chicken groups. Therefore, this experiment is over a new, composite dataset.

\section{Additional Secondary Analysis of MIME}
\label{sec:additional_results}
\begin{figure*}[t]
    \centering
    \includegraphics[width=\textwidth]{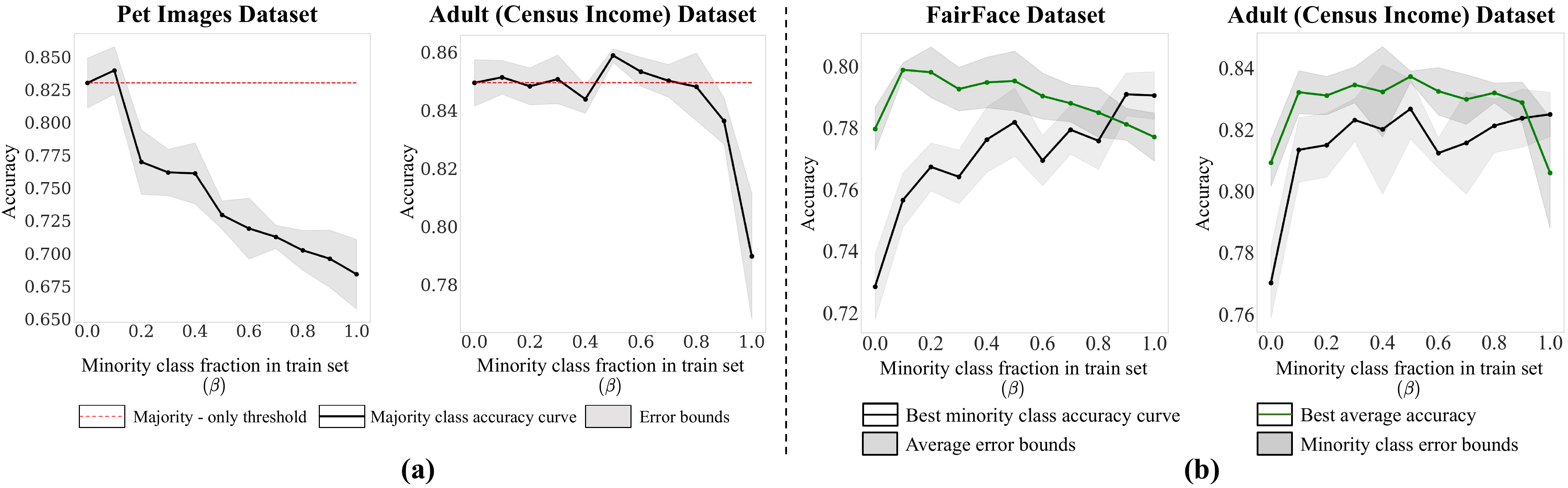}
    \caption{\textbf{The MIME effect is complementary to data debiasing methods and consistent with research aimed at equal representation (ER) datasets.} (a) Training configurations using data debiasing methods~\cite{dong2017class} show the MIME effect. (b) While ER datasets are not optimal for the MIME effect (\textcolor{blue}{Figure~5} and \textcolor{blue}{6}, main paper), optimal overall performance is observed close to ER.}
    \label{fig:discussion}
\end{figure*}

\noindent\textbf{MIME effect with debiasing methods:}
We now analyze the interaction of the MIME effect with existing debiasing methods. Specifically, while applying hard-sample mining~\cite{dong2017class} (as an exemplary case) across the task classes ($y=1,2$), we sweep across various minority training ratios. Figure~\ref{fig:discussion}(a) shows results on two datasets (implementation details may be found in the following section). The MIME effect continues to be observed. Debiasing methods act on the task classes ($y=1,2$) in an effort to improve performance while MIME acts on majority and minority groups, regardless of the task class. Therefore, MIME is complementary to debiasing methods, rather than a competitor. In our experiments, hard-sample mining does not lead to significant performance gains since the task classes are balanced by experimental design. In other scenarios where this might not be the case, MIME and hard sample mining might together improve performance.

\noindent\textbf{Reconciling MIME with existing equal representation (ER) datasets:} In this paper, we focus only on majority group performance, for which ER training datasets are not optimal in general. In contrast, existing efforts~\cite{gebru2018datasheets,buolamwini2018gender,larrazabal2020gender,ryu2017inclusivefacenet,li2019repair,mehrabi2021survey,jo2020lessons,gong2019diversity,kadambi2021achieving} focus on ER datasets to maximize overall (majority+minority) performance. This need not be optimal but is a good thumb rule. This is because while majority group performance eventually reduces with minority training ratio, minority group performance increases (Figure~\ref{fig:discussion}(b) highlights this).

\section{Hard Mining Baseline Implementation}
\label{sec:hardmining}
We implement a version of the method proposed in~\cite{dong2017class}. From a batch of 30 samples, 12 samples (6 of each task class) are retained and used in the training step. These are the samples with least confidence, with respect to ground truth targets. Code is shown below. Trial random seeds are the same as shown in Table~\ref{tab:seeds}.
\begin{python}
class compute_crossentropyloss_hardMine:
    """
    y0 is the vector with shape (batch_size,C)
    x shape is the same (batch_size), whose entries are integers from 0 to C-1
    In our case, C=2.
    """
    def __init__(self, ignore_index=-100) -> None:
        self.ignore_index=ignore_index
    
    def __call__(self, y0, x):
        loss = 0.
        eps = 1e-5
        K = 6
        n_batch, n_class = y0.shape
        pos_score = torch.ones(n_batch).to(device)
        neg_score = torch.ones(n_batch).to(device)
        ix_pos = 0
        ix_neg = 0
        for y1, x1 in zip(y0, x):
            class_index = int(x1.item())
            score = torch.exp(y1[class_index])/(torch.exp(y1).sum()+eps)
            if class_index == 0:
                neg_score[ix_neg] = score
                ix_neg+=1
            else:
                pos_score[ix_neg] = score
                ix_pos+=1
        
        pos_score,_ = torch.sort(pos_score,dim=0)
        neg_score,_ = torch.sort(neg_score,dim=0)
        pos_els = np.minimum(K,ix_pos)
        neg_els = np.minimum(K,ix_neg)
        for ix in np.arange(pos_els):
            loss = loss -torch.log(pos_score[ix])

        for ix in np.arange(neg_els):
            loss = loss -torch.log(neg_score[ix])

        loss = loss/(pos_els+neg_els)
        torch.cuda.empty_cache()
        return loss
\end{python}

\section{Our Code}
\label{sec:code}
\noindent Our code may be accessed through the project webpage at \url{https://visual.ee.ucla.edu/mime.htm/}. We provide code and guidance to perform experiments on all six datasets. Due to specific requirements for each dataset, we provide six Jupyter notebooks. We also include details on setting up file structures and link to datasets wherever necessary. Please refer to the README file for further details.

\section{Negative Impacts and Mitigation}
\label{sec:negative}
\noindent This paper focuses on highlighting the existence of the MIME effect, and not optimal configurations for performance gain. Nevertheless, potential negative outcomes may occur if the results are misinterpreted as guidance on dataset construction with respect to certain stakeholder groups. The rigor of our theoretical results emphasizes this nuance to computer scientists, and future work in diverse venues can extend the notion of minority inclusion for majority group performance gains to broader audiences.

\clearpage
%
%
\bibliographystyle{splncs04}
\bibliography{egbib}